\documentclass{article}

\usepackage[numbers]{natbib}
\usepackage[preprint]{neurips_2025}

\usepackage[utf8]{inputenc} 
\usepackage[T1]{fontenc}    
\usepackage{hyperref}       
\usepackage{url}            
\usepackage{booktabs}       
\usepackage{amsfonts}       
\usepackage{nicefrac}       
\usepackage{microtype}      
\usepackage{xcolor}         
\usepackage{latexsym}
\usepackage{booktabs}
\usepackage{adjustbox}
\usepackage{multirow}
\usepackage[T1]{fontenc}
\usepackage[utf8]{inputenc}
\usepackage{graphicx}

\usepackage{colortbl}
\usepackage{enumitem}%
\usepackage{caption}
\usepackage{multicol}
\usepackage{wrapfig}
\usepackage{amsmath} 
\usepackage{svg}
\usepackage{float}
\usepackage{tocloft}
\usepackage{caption}
\usepackage{etoc}
\usepackage{titletoc}
\usepackage[utf8]{inputenc}
\usepackage[T1]{fontenc}
\colorlet{shadecolor}{gray!30}

\DeclareUnicodeCharacter{03B2}{\ensuremath{\beta}}
\newcommand{\YSR}[1]{\textcolor{blue}{#1}}

\title{MolVision: Molecular Property Prediction with Vision Language Models}

\author{
 \textbf{Deepan Adak\textsuperscript{1}},
 \textbf{Yogesh Singh Rawat\textsuperscript{2}},
 \textbf{Shruti Vyas\textsuperscript{2}}
\\
 \textsuperscript{1}NIT Kurukshetra,
 \textsuperscript{2}University of Central Florida
}
\makeatletter
\AtBeginDocument{
\let\@oldmaketitle\@maketitle
\renewcommand{\@maketitle}{\@oldmaketitle
  \vspace{-20pt}
\begin{minipage}{0.31\textwidth}
    \includesvg[width=\linewidth]{Radar-Plot-Teaser.svg}
\end{minipage}%
\begin{minipage}{0.32\textwidth}
    \includesvg[width=\linewidth]{Radar-plot-regression-Teaser.svg}
\end{minipage}
\begin{minipage}{0.29\textwidth}
    \includesvg[width=\linewidth]{Radar-Plot-Janus-Teaser.svg}
\end{minipage}
  \vspace{-15pt}
  \captionof{figure}{
    \textbf{\textit{MolVision overview:}} Average performance comparison of models in zero-shot (ZS), in-context (ICL), chain-of-thoughts (CoT), and finetuning (FT) for classification (\textbf{\textit{Left} $\uparrow$}) and regression tasks (\textbf{\textit{Center} $\downarrow$}). (\textbf{\textit{Right:)}} Impact of using visual information on model performance ($\uparrow$) (JanusPro).
  }
  \label{fig:teaser}
 }
}
\makeatother
\begin{document}

\maketitle

\begin{abstract}
  Molecular property prediction is a fundamental task in computational chemistry with critical applications in drug discovery and materials science. While recent works have explored Large Language Models (LLMs) for this task, they primarily rely on textual molecular representations such as SMILES/SELFIES, which can be ambiguous and structurally less informative. In this work, we introduce MolVision, a novel approach that leverages Vision-Language Models (VLMs) by integrating both molecular structure as images and textual descriptions to enhance property prediction. We construct a benchmark spanning ten diverse datasets, covering classification, regression and description tasks. Evaluating nine different VLMs in zero-shot, few-shot, and fine-tuned settings, we find that visual information improves prediction performance, particularly when combined with efficient fine-tuning strategies such as LoRA. Our results reveal that while visual information alone is insufficient, multimodal fusion significantly enhances generalization across molecular properties. Adaptation of vision encoder for molecular images in conjunction with LoRA further improves the performance. The code and data is available at : \href{https://molvision.github.io/MolVision/}{https://molvision.github.io/MolVision/}.
\end{abstract}

\section{Introduction}\label{sec:introduction}

Recent advancements in Large Language Models (LLMs) have revolutionized natural language understanding and generation across multiple domains \cite{openai2024gpt4}. Models such as GPT \cite{achiam2023gpt, floridi2020gpt}, LLaMA \cite{touvron2023llama}, and Mistral \cite{jiang2023mistral} have demonstrated exceptional capabilities in reasoning, knowledge retrieval, and complex problem-solving. Extending beyond pure text-based reasoning, Vision-Language Models (VLMs) integrate visual and textual modalities \cite{openai2024gpt4, liu2023visual, gao2023llamaadapter, geminiteam2024gemini}, enabling them to perform tasks such as image captioning, visual question answering, and multimodal retrieval with remarkable success. While VLMs have been extensively explored in computer vision and NLP applications, their potential in scientific domains—particularly in chemistry—remains largely unexplored. Given that molecular structures are inherently visual, leveraging vision in molecular analysis presents an exciting, yet underexplored, research direction.

Recent works such as ChemLLM \cite{zhang2024chemllm} and ChemLLM-Bench \cite{guo2023large} have begun to explore LLMs for molecular property prediction. These methods primarily rely on textual molecular representations, such as SMILES and SELFIES, which have been widely used in cheminformatics for decades. However, these representations have notable limitations, including their non-uniqueness and syntactic instability, where structurally identical molecules may have vastly different textual encodings. This ambiguity introduces challenges for LLMs, which process molecular structures as linear strings, potentially overlooking key structural relationships. While some approaches attempt to improve these representations through graph-based models \cite{lu2023highly}, the integration of easily available visual molecular data remains largely unexplored in this domain.

Incorporating visual information has the potential to significantly enhance molecular property prediction. Chemists usually analyze molecular structures using bond-line or skeletal diagrams to infer properties such as reactivity, toxicity, and solubility. These visual representations inherently encode structural and spatial information that textual descriptors may fail to capture. For example, subtle differences in geometry, stereochemistry, or electron delocalization can have profound effects on molecular properties, yet are difficult to represent accurately in SMILES format alone. By leveraging VLMs, which are designed to process both visual and textual inputs, we aim to bridge this gap and improve predictive modeling in cheminformatics.

\begin{figure}[t!]
\centering
\includegraphics[width=0.95\textwidth]{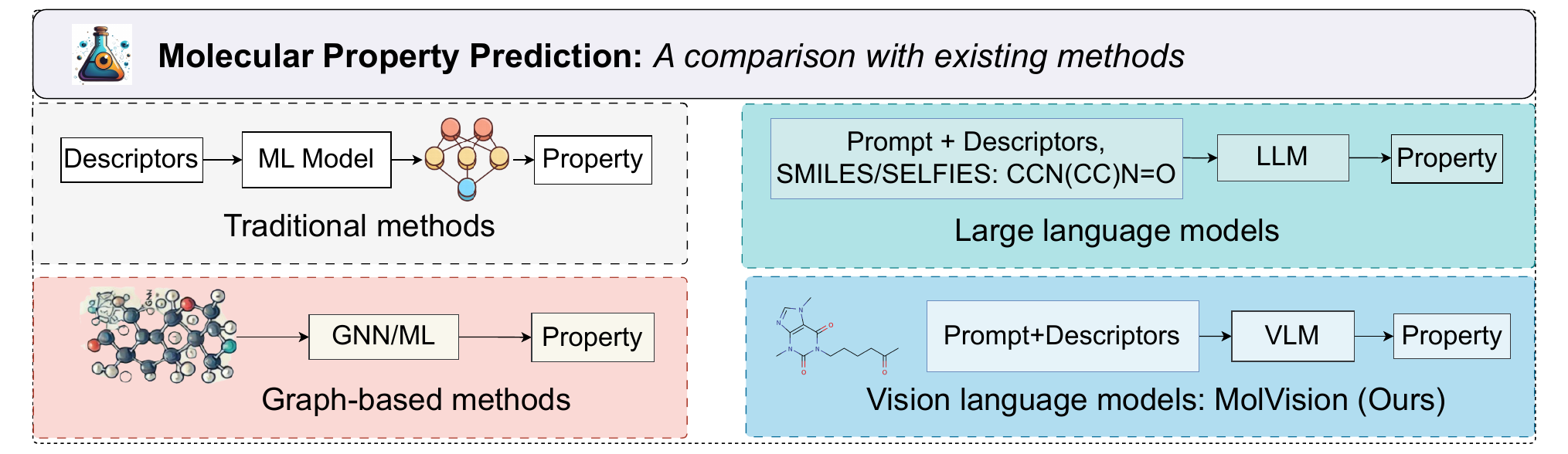}
\vspace{-5pt}
\caption{\textbf{\textit{MolVision comparison:}} Comparison of relevant molecular property prediction approaches.
}
\label{fig:teaser_2}
\vspace{-15pt}
\end{figure}

To this end, we introduce MolVision, a multimodal benchmark for molecular property prediction. In contrast to prior works MolVision integrates both textual and visual representations (Figure \ref{fig:teaser_2}). Our benchmark spans ten diverse datasets, covering classification, regression and description tasks across a wide range of molecular properties, including toxicity, solubility, and bioactivity. We evaluate nine different VLMs in zero-shot, few-shot, and fine-tuned settings, providing a comprehensive analysis of their performance in this domain (Figure \ref{fig:teaser}). 
We also propose a simple contrastive strategy to adapt visual component of VLMs for this domain and demonstrate its effectiveness for property prediction.

Through extensive experimentation, we uncover several key insights. First, while VLMs struggle in zero-shot settings, their performance improves significantly with in-context learning and fine-tuning. Second, efficient adaptation techniques such as LoRA enhance the predictive accuracy and generalization to unseen molecular properties. Third, while visual information alone is insufficient for accurate property prediction, combining molecular images with textual representations yields notable performance gains specifically for larger molecules.
These findings suggest that vision-augmented molecular modeling presents a promising avenue for future research in AI-driven chemistry, with numerous potential applications. 
We make the following contributions:
\begin{itemize}[topsep=2pt]
\setlength\itemsep{-0.2em}
    \item We introduce MolVision, a novel approach for molecular property prediction, integrating molecular structure images with textual representations. 
    \item We present a multimodal benchmark and systematically assess nine state-of-the-art VLMs in zero-shot, few-shot, and fine-tuned settings across ten datasets highlighting their strengths and limitations for property prediction.
   \item We show that efficient adaptation of VLMs for property prediction enhances both performance and generalization, and that combining visual and textual data significantly improves molecular property prediction. 
   \item We propose a simple contrastive strategy, implemented with LoRA, to efficiently adapt vision aspect of VLMs for this domain and demonstrate it effectiveness for property prediction.
\end{itemize}

\begin{figure*}[t!]
\centerline{
    \includegraphics[width=0.95\textwidth]{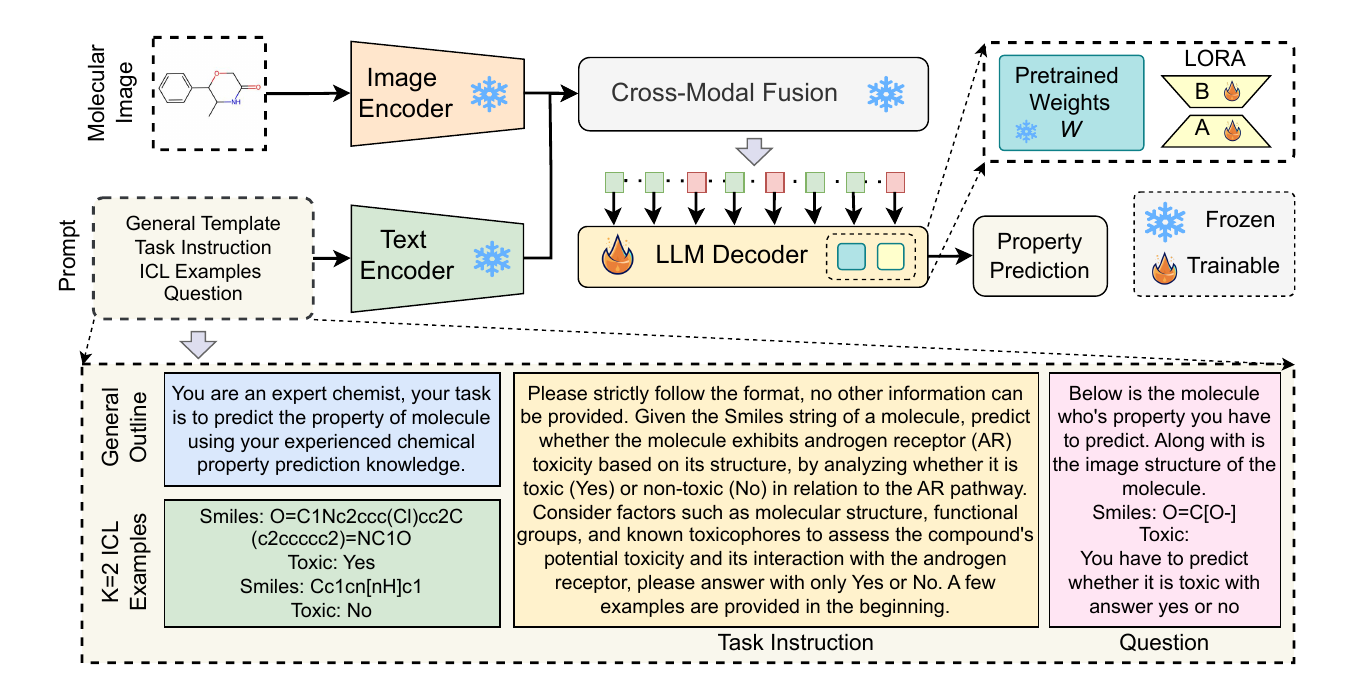}
}
\caption{\textbf{\textit{Overview of visual-textual approach for property prediction:}} The image representation along with textual description are used as input by the VLM where the image is encoded by a vision encoder and textual description is encoded by a text encoder. These multimodal features are used to generate the output with the help of a decoder. We show template prompt used for property prediction, including general outline, task instruction, in-context learning (ICL with k=2), and an image prompt. 
}
\label{fig:flowchartprompt}
\vspace{-10pt}
\end{figure*}

\vspace{-10pt}
\section{Related works}
\vspace{-5pt}
\noindent
\textbf{Property prediction:}
Prior research in molecular property prediction has explored various methods and representations \cite{Hasselgren_2024}. Traditional approaches, like molecular fingerprints \cite{tazhigulov2022molecular,kumar2023coarsegrained} and descriptors \cite{todeschini2008handbook}, rely on expert knowledge but are limited in capturing complex data relationships. Recently, machine learning techniques \cite{Choudhary_2023,_zt_rk_2020,Hocky_2022}, particularly graph-based methods like graph convolutional networks (GCNs) \cite{gilmer2017neural}, have gained prominence for capturing molecular interactions. Additionally, deep learning models, including RNNs, CNNs, and transformers \cite{wu2018moleculenet,doi:10.1021/ci00057a005,tang2020self,goh2017deep,elton2019deep}, have shown strong performance in modeling structural and sequential information from molecular data.


\noindent
\textbf{Multimodal foundational models:}
Recent advances in Foundational Models \cite{jiang2023mistral,touvron2023llama,brown2020language} have shown the ability of multimodal LLMs to process both vision and language. These models integrate vision encoders \cite{fang2022eva,radford2021learning} with LLMs \cite{zheng2023judging,touvron2023llama} for generating text responses. Models like Llama Adapter V2 \cite{gao2023llamaadapter} and Flamingo \cite{alayrac2022flamingo} explored multimodal structures. Typically, these models pre-train on image caption datasets \cite{changpinyo2021conceptual,lin2015microsoft} and fine-tune on task-specific datasets \cite{agrawal2016vqa}. Models such as Llava 1.5 \cite{liu2024improved} and QwenVL \cite{bai2023qwenvl} are designed for instruction-following tasks but may struggle with science-specific challenges like computational chemistry.

\noindent
\textbf{Foundation models for property prediction:}
Recent efforts have explored LLMs for property prediction, such as ChemLLM \cite{zhang2024chemllm}, ChemLLMBench \cite{guo2023large}, and Nach0 \cite{livne2024nach0}. ChemLLM \cite{zhang2024chemllm} utilizes ChemData, an instruction-tuning dataset, to address the need for specialized models in chemistry. Guo et al. \cite{guo2023large} assess LLMs in chemistry, focusing on understanding and reasoning tasks with zero-shot and few-shot learning. In \cite{liu2023molca}, the authors propose to utilize graphical structure with LLMs for molecule captioning, IUPAC name prediction, and molecule-text retrieval. In contrast, our work examines the role of multimodal vision-language models, incorporating visual and textual data for molecular property prediction, a first-of-its-kind exploration.

\section{Visual language models for property prediction}

We propose use of visual information, in the form of molecular images, alongside textual descriptions to improve property prediction. Images provide structural insights that are challenging to interpret from text alone. A vision-language model processes both the image and text prompt to generate a textual output, as shown in Figure \ref{fig:flowchartprompt}.
The input image is divided into patches, which are converted into tokens for the vision encoder (e.g., ViT \cite{li2023blip2}). The textual prompt is passed through a text encoder (e.g., BERT \cite{li2023blip2}), and the visual and textual features are fused via multi-modal learning (e.g., using Q-Former). LLM decoder (e.g., a transformer model) uses these fused features for text generation.



\noindent \textbf{Text prompt:} The prompt consists of three components passed to the text encoder: 1) \textit{General outline} provides an overview of the task, 2) \textit{Task instruction} includes detailed task-specific guidance, and 3) \textit{Question} requests the model’s answer in a specific format. For in-context learning and chain-of-thoughts, additional information like examples or reasoning steps is included (Figure \ref{fig:flowchartprompt}). 



\subsection{Model variants}


We study three different setups: 1) zero-shot, 2) few-shot, and 3) fine-tuning on training data. 


\begin{figure*}[t!]
\centerline{\includegraphics[width=.95\textwidth]{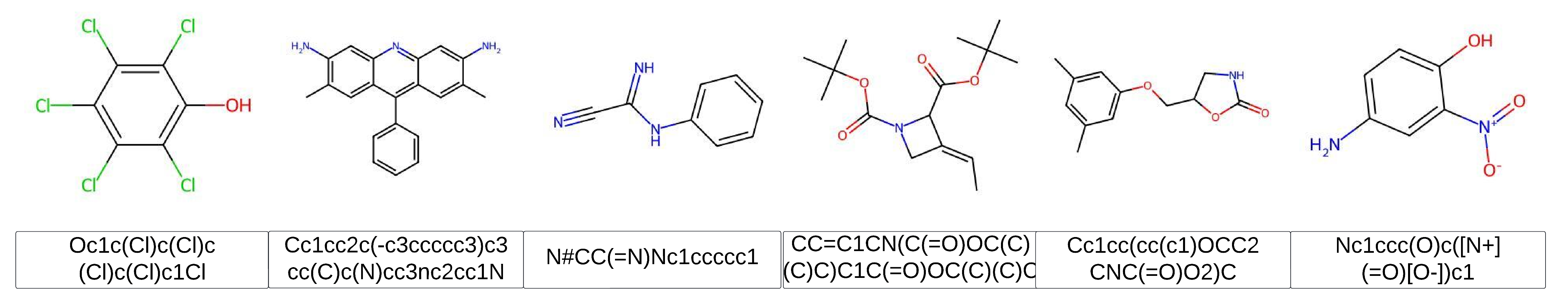}}
\vspace{-2pt}
\caption{\textbf{\textit{Sample visual and textual representation pairs:}} The images in top row shows skeletal structure of molecules and bottom row shows their corresponding SMILES representations.
}
\label{fig:structure}
\vspace{-12pt}
\end{figure*}

\noindent
\textbf{Zero-shot:} The model is evaluated without fine-tuning or in-context examples.

\noindent
\textbf{Few-shot:} 
We use two approaches: 1) \textit{In-context Learning (ICL)}: traditional prompting (labeled ICL) and \textit{Chain of Thought (CoT) Prompting}. Traditional ICL constructs prompts with examples similar to the input, enabling the model to learn relationships in the new domain. CoT prompting enhances reasoning by guiding the model through intermediate steps. In cheminformatics, molecular similarity is often quantified using methods like similarity and distance metrics. For selecting few-shot samples, we use the Tanimoto index, an effective parameter for similarity prediction \cite{bai2023qwenvl}.

\noindent
\textbf{Finetuning:} 
Vision-language models (VLMs) possess a substantial number of trainable parameters, rendering traditional fine-tuning impractical as all model parameters undergo gradient updates simultaneously. In our study, we adopt LoRA (Low-Rank Adaptation) \cite{hu2021lora} for efficient fine-tuning, a technique that significantly reduce the number of trainable parameters. LoRA achieves this by updating weights through a pair of trainable rank decomposition matrices, which operate in parallel with existing weight matrices, while keeping the original pre-trained weights frozen during fine-tuning. We only adapt the LLM decoder keeping other components frozen during this finetuning to preserve the generalization capabilities of vision and text encoders (Figure \ref{fig:flowchartprompt}). 


\subsection{Model architectures}
We study nine different state-of-the-art visual language models in this study. This includes both closed-source and open-source models. In open-source, we experimented with Janus-Pro 7B \cite{chen2025janus}, BLIP-2 \cite{li2023blip2}, Llava 1.5 \cite{liu2024improved}, Llama Adapter V2 \cite{gao2023llamaadapter}, CogVLM \cite{wang2024cogvlm}, Qwen-VL \cite{bai2023qwenvl}, and mPLUGOWL2 \cite{ye2024mplugowl}. For closed-source, we experimented with GPT-4V and GPT-4o \cite{openai2024gpt4}. 

\vspace{-5pt}
\section{MolVision benchmark}\label{sec:benchmark}
\vspace{-5pt}
In this section, we introduce the MolVision benchmark, which includes ten diverse datasets. These datasets cover a wide range of properties, such as molecular weight, topological 
polar surface area, and toxicity, and encompass classification, regression  and description tasks. A summary in Table \ref{tab:molecular_datasets}.



\noindent
\textbf{Tasks:}
VLMs generate textual outputs based on image and text prompts. For classification, we frame the task as a True/False question, where the model predicts whether a molecule inhibits a target property. For regression, the model generates a numerical value representing the target property and for description task, the model generates textual output.

\noindent
\textbf{Dataset curation:}
We incorporate both molecular skeletal structures as images and SMILES/SELFIES representations. Existing property prediction datasets primarily focus on textual representations like SMILES, lacking structural images. To address this, we augment these datasets with skeletal images generated using RDKit \cite{landrum2013rdkit}. Figure \ref{fig:structure} illustrates examples of skeletal (bond-line) structures alongside their SMILES representations. RDKit also enables conversion between SMILES and SELFIES, allowing us to explore diverse molecular encodings and enhance model robustness.
\subsection{Benchmark datasets}
\begin{wraptable}{r}{0.4\textwidth}
\vspace{-10pt}
\caption{\textbf{\textit{MolVision benchmark details}:} Statistics of datasets used in this study.}
\label{tab:molecular_datasets}
\centering
\footnotesize
\setlength{\tabcolsep}{1.5pt}
\begin{tabular}{l|r|r|l}
\toprule
\textbf{Dataset} & \textbf{Train} & \textbf{Test} & \textbf{Property} \\
\midrule
\rowcolor{lightgray} \multicolumn{4}{l}{\textbf{Classification}} \\
\midrule
BACE-V & 1,210 & 303 & Bioactivity \\
BBBP-V & 1,640 & 410 & BBB Pen. \\
HIV-V & 32,902 & 8,225 & HIV activity \\
ClinTox-V & 1,193 & 298 & Toxicity \\
Tox21-V & 6,265 & 1,566 & Toxicity \\
\midrule
\rowcolor{lightgray} \multicolumn{4}{l}{\textbf{Regression}} \\
\midrule
ESOL-V & 902 & 226 & Solubility \\
LD50-V & 5,908 & 1,477 & Toxicity \\
QM9-V & 107K & 27K & Quantum \\
PCQM4Mv2-V & 3.0M & 0.7M & Quantum \\
\midrule
\rowcolor{lightgray} \multicolumn{4}{l}{\textbf{Molecular Description}} \\
\midrule
ChEBI-V & 32,000 & 8,000 & Description \\
\bottomrule
\end{tabular}
\vspace{-10pt}
\end{wraptable}

The curation process provides both image representations corresponding to each molecule along with a formatted prompt which is derived through manual engineering. The benchmark consists of the following datasets (more details in Appendix): \textbf{BACE-V} is derived from BACE (Binary Activity of Chemical Entities) dataset \cite{doi:10.1021/acs.jcim.6b00290} which is widely used for binary classification in bioactivity prediction, particularly for BACE-1 inhibitors linked to Alzheimer's. \textbf{BBBP-V} is based on the Blood-Brain Barrier Penetration (BBBP) dataset \cite{bbbp}, which provides binary labels for BBB penetration. \textbf{HIV-V} is based on the HIV \cite{hiv} where we focus on predicting HIV replication inhibition. \textbf{Clintox-V} is derived from ClinTox dataset \cite{doi:10.1021/acs.jcim.6b00290} and our focus is on predictions of clinical toxicity and FDA approval status. \textbf{Tox21-V} is based on the Tox21 dataset \cite{huang2016tox21challenge} and focuses on predicting chemical toxicity, critical for environmental safety. \textbf{ESOL-V} is based on the ESOL dataset \cite{delaney2004esol}, and focus on predicting aqueous solubility of organic compounds. \textbf{LD50-V} is based on the LD50 \cite{karmaus2018variability} and focuses on acute toxicity. \textbf{QM9-V} is derived from the QM9 \cite{jain2013commentary} and focuses on quantum chemical properties. \textbf{PCQM4Mv2-V} is derived from the PCQM4Mv2 dataset \cite{hu2021ogb} and focuses on predicting the HOMO-LUMO gap. \textbf{ChEBI20-V} is derived from the Chemical Entities of Biological Interest (ChEBI) \cite{edwards-etal-2021-text2mol}database and focuses on generating accurate textual descriptions of molecular structures.

\vspace{-5pt}
\section{Experiments and results}
\vspace{-5pt}
Next, we provide evaluations on MolVision benchmark followed by some discussion and analysis.

\noindent
\textbf{Evaluation metrics:}
We evaluate classification performance using Accuracy and F1 Score. 
We evaluate classification using Accuracy and F1 Score, where Accuracy measures correct predictions, and F1 Score balances precision and recall. For regression, we use mean absolute error (MAE) and root mean square error (RMSE) to quantify prediction deviations. Molecular description tasks are assessed with BLEU-2, BLEU-4, ROUGE-1, ROUGE-2, ROUGE-L, and METEOR, capturing n-gram precision, sequence overlap, and semantic similarity.

\subsection{Benchmarking results}
All experiments are conducted with a temperature of 0 (unless stated) to reduce prediction volatility.

\noindent
\textbf{Zero-shot:} 
VLMs are trained on large-scale datasets to learn associations between visual and textual features. However, property prediction presents a distinct challenge, differing from their training domain. Figures \ref{fig:teaser} (left and center) show zero-shot results across all datasets, where performance remains low for most models, except for proprietary models GPT-4o and GPT-4v. (More in Appendix.)

\noindent
\textbf{Few-shot:} 
Tables \ref{table:performance_ICL} and \ref{table:performance_ICL_reg} present few-shot ICL performance for classification and regression tasks, respectively. All models show performance gains over zero-shot, though Llama Adapter v2 7B and Qwen VL consistently underperform with classification accuracy below 48\% and high regression errors. BBBP-V and QM9 remain challenging datasets, while Tox21-V and LD50 yield comparatively better results. As expected, closed models such as GPT-4o and GPT-4v achieved the best performance across most datasets however Janus-Pro 7B performed better on certain datasets. The performance was followed by Llava 1.5 13B and BLIP-2 as the second best open-source models for classification and regression, respectively. 
Table \ref{table:performance_ICL_reg} also reports CoT prompting results for regression, showing further improvements, also seen in classification tasks (Figure \ref{fig:teaser}, more info in Appendix).

\begin{table*}[t!]
    \caption{\textbf{\textit{Few-shot performance for classification tasks:}} A comparison of accuracy (f1-score) on property prediction task using in-context learning (ICL with k=2). ($\dagger$ - fully supervised training)
    }
    \label{table:performance_ICL}
    \vspace{-5pt}
    \centering \footnotesize
    \begin{adjustbox}{max width=0.85\textwidth,center}
    \begin{tabular}{lccccc|c}
        \toprule
        \textbf{Models} & \textbf{BACE-V $\uparrow$} & \textbf{BBBP-V$\uparrow$} & \textbf{HIV-V $\uparrow$} & \textbf{ClinTox-V $\uparrow$} & \textbf{Tox21-V $\uparrow$} & \textbf{Average $\uparrow$} \\
         \midrule
                  \rowcolor{shadecolor} 
        \multicolumn{7}{l}{GNN Models} \\
        \midrule
                UniMol \cite{lu2023highly} $\dagger$ &0.78(0.67)&0.82(0.70)&	0.82(0.73)&	0.94(0.83)&	0.77(0.65) &0.83(0.72) \\
                Molca \cite{liu2023molca} $\dagger$ & 0.79(0.73)	&0.74(0.72)	&0.89(0.84)&	0.93(0.84)	&0.80(0.72)&
                0.83(0.77)\\
                
         \rowcolor{shadecolor} 
         \midrule
        \multicolumn{7}{l}{LLM [ICL k=2]} \\
        \midrule
        Guo et. al. \cite{guo2023can}& {0.49(0.40)} & {0.46(0.46)} & {0.86}(0.80) & {0.57(0.36)} & {0.57}(0.52) & 0.59(0.51) \\
        Davinci-003 &0.65(0.64) & 0.39(0.37)&0.78(0.83)& 0.84(0.85)&0.68(0.51) & 0.67(0.64) \\
        ChemLLM\cite{zhang2024chemllm}& 0.18(0.12) & 0.12(0.08) & 0.19(0.09) & 0.21(0.13) &0.18(0.09)& 0.18(0.10) \\
        Gal-1.3B \cite{taylor2022galactica} & 0.38(0.29) & 0.42(0.26) & 0.33(0.23) & 0.40(0.34) & 0.49(0.32) & 0.40(0.29)\\
                Gal-6.7B \cite{taylor2022galactica} & 0.41(0.30) & 0.44(0.28) & 0.35(0.25) & 0.43(0.36) & 0.52(0.34) & 0.44(0.31) \\
        \midrule
        \rowcolor{shadecolor} 
        \multicolumn{7}{l}{VLM [ICL k=2]} \\
        \midrule
        GPT-4o & 0.56(0.53) & \textbf{0.77(0.81)} & 0.82(\textbf{0.56}) & 0.59(0.44) & 0.42(0.58) & 0.63\underline{(0.58)} \\
        GPT-4v & 0.72(0.66) & 0.63(0.60) & \textbf{0.95}(0.44) & \textbf{0.96}(\textbf{0.94}) & 0.72(0.52) & \textbf{0.80(0.63)} \\
        Janus Pro 7B & \textbf{0.78(0.71)} & 0.68(0.62) & 0.92(0.52) & 0.83(0.56) & 0.69(0.49) & \underline{0.78}(0.58) \\
        BLIP-2 & 0.36(0.52) & 0.37(0.29) & 0.60{(0.30)} & 0.34(0.36) & 0.75(0.42) & 0.48(0.38) \\
        Llava 1.5 13B & 0.49(0.48) & 0.44(0.39) & 0.24(0.34) & 0.64(0.76) & \textbf{0.81}(0.31) & 0.52(0.46) \\
        Llama Ad v2 7B & 0.28(0.29) & 0.18(0.11) & 0.19(0.17) & 0.29(0.12) & 0.31(0.21) & 0.25(0.18) \\
        CogVLM & 0.48(0.51) & 0.40(0.37) & 0.31(0.21) & 0.64(0.62) & 0.69\textbf{(0.65)} & 0.50(0.47) \\
        QwenVL & 0.69(0.46) & 0.30(0.12) & 0.28(0.36) & 0.52(0.48) & 0.62(0.63) & 0.48(0.41) \\
        mPlugowl2 & 0.59(0.32) & 0.35(0.38) & 0.62(0.29) & 0.34(0.42) & 0.69(0.56) & 0.52(0.39) \\
        \bottomrule
    \end{tabular}
    \end{adjustbox}
    \vspace{-1pt}
\end{table*}

\begin{table*}[t!]
    \caption{\textbf{\textit{Few-shot performance for regression:}} A comparison of error in prediction using in-context learning (ICL k=2) and chain-of-thoughts (CoT) with traditional and LLM based approaches. 
    }
    \label{table:performance_ICL_reg}
    \centering \footnotesize
   \begin{adjustbox}{max width=0.9\textwidth}
    \begin{tabular}{@{}l|cc|cc|cc|cc|cc@{}}
    \toprule
        {\textbf{Model}} & \multicolumn{2}{c|}{\textbf{ESOL-V} $\downarrow$} & \multicolumn{2}{c|}{\textbf{LD50-V $\downarrow$}} & \multicolumn{2}{c|}{\textbf{QM9-V $\downarrow$}} & \multicolumn{2}{c|}{\textbf{PCQM4M-V $\downarrow$}} & \multicolumn{2}{c}{\textbf{Average $\downarrow$}} \\
        \midrule
        \rowcolor{shadecolor} 
        \multicolumn{11}{l}{Traditional approaches} \\
        \midrule
        GenRA\cite{helman2019transitioning} & \multicolumn{2}{c|}{-} & \multicolumn{2}{c|}{0.58} & \multicolumn{2}{c|}{-} & \multicolumn{2}{c|}{-} & \multicolumn{2}{c}{-} \\
        Unimol \cite{lu2023highly} & \multicolumn{2}{c|}{0.788} & \multicolumn{2}{c|}{-} & \multicolumn{2}{c|}{0.00467} & \multicolumn{2}{c|}{0.070} & \multicolumn{2}{c}{-} \\
        \midrule
        \rowcolor{shadecolor} 
        \multicolumn{11}{l}{Large Language Models} \\
        \midrule
        GPT-3.5 & \multicolumn{2}{c|}{4.24} & \multicolumn{2}{c|}{11.67} & \multicolumn{2}{c|}{13.52} & \multicolumn{2}{c|}{1.81} & \multicolumn{2}{c}{5.81} \\
        Llama2 13B & \multicolumn{2}{c|}{27.71} & \multicolumn{2}{c|}{49.22} & \multicolumn{2}{c|}{78.92} & \multicolumn{2}{c|}{102.92} & \multicolumn{2}{c}{64.69} \\
        Mistral 13B & \multicolumn{2}{c|}{33.21} & \multicolumn{2}{c|}{27.46} & \multicolumn{2}{c|}{66.80} & \multicolumn{2}{c|}{88.90} & \multicolumn{2}{c}{54.09} \\
        ChemLLM \cite{zhang2024chemllm} & \multicolumn{2}{c|}{23.42} & \multicolumn{2}{c|}{33.91} & \multicolumn{2}{c|}{147.10} & \multicolumn{2}{c|}{29.01} & \multicolumn{2}{c}{58.36} \\
        GAL-1.3B \cite{taylor2022galactica} & \multicolumn{2}{c|}{18.92} & \multicolumn{2}{c|}{40.49} & \multicolumn{2}{c|}{140.92} & \multicolumn{2}{c|}{29.92} & \multicolumn{2}{c}{-} \\
        Gal-6.7B \cite{taylor2022galactica}& \multicolumn{2}{c|}{13.47} & \multicolumn{2}{c|}{38.02} & \multicolumn{2}{c|}{128.90} & \multicolumn{2}{c|}{28.48} & \multicolumn{2}{c}{-} \\
        Molca \cite{liu2023molca}  & \multicolumn{2}{c|}{1.849} & \multicolumn{2}{c|}{0.982} & \multicolumn{2}{c|}{4.889} & \multicolumn{2}{c|}{0.802} & \multicolumn{2}{c}{-} \\
       \midrule
\rowcolor{shadecolor} 
\multicolumn{1}{l}{Vision-Language Models} & ICL & CoT & ICL & CoT & ICL & CoT & ICL & CoT & ICL & CoT \\
\midrule
GPT-4o & 0.98 & 0.77 & 0.87 & 0.60 & \textbf{8.38} & 5.24 & 0.68 & 0.53 & 2.73 & 1.78 \\
GPT-4v & 0.99 & \textbf{0.71} & \textbf{0.71} & 0.59 & 8.62 & 4.66 & 0.77 & 0.66 & 2.78 & 1.66 \\
Janus-Pro 7B & \textbf{0.61} & 0.89 & 0.72 & 0.60 & 8.53 & \textbf{4.42} & \textbf{0.62} & \textbf{0.38} & \textbf{2.52} & \textbf{1.57} \\
BLIP-2 & 1.99 & 1.07 & 0.73 & \textbf{0.49} & 16.01 & 10.09 & 1.30 & 1.25 & 5.01 & 3.23 \\
Llava 1.5 13B & 6.01 & 2.18 & 0.94 & 0.69 & 27.00 & 15.21 & 1.42 & 1.49 & 8.84 & 4.89 \\
Llama Ad v2 7B & 3.08 & 2.17 & 3.36 & 2.12 & 28.09 & 19.24 & 4.06 & 2.36 & 9.15 & 6.47 \\
CogVLM & 1.26 & 1.21 & 3.47 & 0.78 & 25.85 & 15.15 & 1.44 & 1.24 & 8.50 & 4.59 \\
Qwen VL & 3.96 & 2.89 & 1.06 & 0.63 & 38.92 & 18.08 & 10.61 & 9.56 & 13.64 & 7.29 \\
mPlugOWL2 & 1.46 & 1.50 & 0.94 & 0.71 & 29.33 & 19.17 & 1.84 & 1.62 & 8.89 & 5.25 \\
    \bottomrule
    \end{tabular}
    \end{adjustbox}
    \vspace{-5pt}
\end{table*}

\begin{table*}[t!]
    \caption{\textbf{\textit{Classification performance after finetuning:}} Accuracy (F1 score) comparison of models finetuned using LoRA. The best performing models are highlighted with bold text. 
    }
    \label{table:performance_lora}
    \centering \footnotesize
    \begin{adjustbox}{max width=0.9\textwidth,center}
    \begin{tabular}{lccccccc}
        \toprule
        \textbf{Models} & \textbf{BACE-V $\uparrow$} & \textbf{BBBP-V$\uparrow$} & \textbf{HIV-V $\uparrow$} & \textbf{ClinTox-V $\uparrow$} & \textbf{Tox21-V$\uparrow$}& \textbf{Average $\uparrow$} \\
         \midrule
        RF & 0.79(0.76) & 0.82(0.88) & 0.87(0.52) & 0.85(0.46) & 0.83(0.26) &0.83(0.57) \\ 
        XGBoost & {0.81(0.77)} & {0.85(0.90)} & {0.87(0.55)} & {0.88(0.62)} & {0.84(0.33)} & 0.85(0.63) \\ 
        ChemLLM \cite{zhang2024chemllm}& 0.18(0.12) & 0.12(0.08) & 0.19(0.09) & 0.21(0.13) &0.18(0.09)& 0.17(0.10) \\
        Molca \cite{liu2023molca} & 0.79(0.73)	&0.74(0.72)	&0.89(0.84)&	0.93(0.84)	&0.80(0.72)&
                0.83(0.77)\\
        \midrule
        BLIP-2 & \textbf{0.86(0.83)} & \textbf{0.93(0.96)} & \textbf{0.92}(0.76) & \textbf{0.89(0.93)} & \textbf{0.99}(0.80) & \textbf{0.92}({0.86}) \\
        Llava 1.5 13B & 0.84(\textbf{0.83}) & 0.86(0.88) & 0.80(0.81) & 0.70(0.72) & 0.92(0.93) & 0.82(0.83) \\
        Llama Adapter v2 7B & 0.52(0.48) & 0.45(0.46) & 0.43(0.42) & 0.58(0.62) & 0.68(0.69) & 0.53(0.53) \\
        CogVLM & 0.72(0.71) & 0.78(0.82) & 0.85(0.83) & 0.88(0.90) & 0.93(0.93) & 0.83(0.84) \\
        Qwen VL & 0.78(0.78) & 0.70(0.72) & 0.60(0.61) & 0.71(0.64) & 0.75(0.64) & 0.71(0.68) \\
        mPlugOWL2 & \textbf{0.86}(0.82) & 0.90(0.88) & 0.90\textbf{(0.91)} & \textbf{0.89}(0.92) & {0.94\textbf{(0.96)}} & {0.89}(\textbf{0.89}) \\
        \bottomrule
    \end{tabular}
    \end{adjustbox}
    \vspace{-5pt}
\end{table*}
\begin{table*}[t!]
    \caption{\textbf{\textit{Molecular description performance after finetuning:}} Comparison of models finetuned using LoRA on the ChEBI dataset. The best performing models are highlighted with bold text.}
    \label{table:performance_chebi_lora}
    \centering \footnotesize
    \begin{adjustbox}{max width=\textwidth,center}
    \begin{tabular}{lccccccc}
        \toprule
        \textbf{Models} & \textbf{BLEU-2 $\uparrow$} & \textbf{BLEU-4 $\uparrow$} & \textbf{ROUGE-1 $\uparrow$} & \textbf{ROUGE-2 $\uparrow$} & \textbf{ROUGE-L $\uparrow$} & \textbf{METEOR $\uparrow$} & \textbf{Average $\uparrow$} \\
        \midrule
        MolT5 \cite{edwards2022translationmoleculesnaturallanguage} & 59.40 & 50.80 & 65.40 & 51.00 & 59.40 & 61.40 & 57.90\\
        Molca \cite{liu2023molca} & 62.00 & 53.10 & 68.10 & 53.70 & 61.80 & 65.10 & 60.60 \\
        \midrule
        BLIP-2 & 59.06 & 58.03 & 58.93 & 58.47 & 58.89 & 58.19 & 58.60 \\
        CogVLM & \textbf{63.00} & \textbf{60.01} & \textbf{62.39} & \textbf{61.16} & \textbf{62.00} & \textbf{60.60} & \textbf{61.52} \\
        mPlugOWL2 & 51.93 & 49.64 & 51.56 & 50.67 & 51.33 & 50.06 & 50.87 \\
        Llava 1.5 13B & 60.88 & 58.99 & 60.62 & 59.80 & 60.42 & 59.40 & 60.02 \\
        Llama Adapter v2 7B & 46.60 & 44.65 & 46.27 & 45.50 & 46.07 & 45.00 & 45.68 \\
        Qwen VL & 52.00 & 50.04 & 51.63 & 50.78 & 51.40 & 50.44 & 51.05 \\
        \bottomrule
    \end{tabular}
    \end{adjustbox}
\end{table*}

\begin{table*}[t!]
\centering
\caption{\textbf{\textit{Impact of visual information:}} 
A performance comparison showing the impact of visual information (molecular image) when used with textual description (SMILES). Accuracy is shown for classification (BACE-V, BBBP-V, HIV-V, Clintox-V (CV), Tox21-V (TV)), and MAE (LD50-V, QM9-V and PCQM4Mv2-V (PV)) and RMSE (ESOL-V) is shown for regression tasks.
}
\label{tab:ablation_multi}
\vspace{0pt}
\centering \footnotesize
\begin{adjustbox}{max width=\textwidth,center}
\begin{tabular}{l|l|*{9}{c}}
\hline
\textbf{Model} & \textbf{Input} & \rotatebox{0}{\textbf{BACE-V $\uparrow$}} & \rotatebox{0}{\textbf{BBBP-V  $\uparrow$}} & \rotatebox{0}{\textbf{HIV-V  $\uparrow$}} & \rotatebox{0}{\textbf{CV  $\uparrow$}} & \rotatebox{0}{\textbf{TV  $\uparrow$}} & \rotatebox{0}{\textbf{ESOL-V  $\downarrow$}} & \rotatebox{0}{\textbf{LD50-V $\downarrow$}} & \rotatebox{0}{\textbf{QM9-V $\downarrow$}} & \rotatebox{0}{\textbf{PV $\downarrow$}} \\
\hline
\multirow{3}{*}{BLIP-2} 
& Text Only & 0.71 & 0.76 & 0.69 & 0.64 & 0.78 & 9.89 & 7.80 & 31.76 & 11.31 \\
& Image Only & 0.15 & 0.09 & 0.10 & 0.13 & 0.18 & 32.16 & 31.23 & 149.12 & 36.96 \\
\rowcolor{shadecolor} 
\cellcolor{white}& Image+Text & 0.86 & 0.93 & 0.92 & 0.89 & 0.99 & 1.07 & 0.49 & 4.92 & 1.99 \\
\hline
\multirow{3}{*}{JanusPro} 
& Text Only & 0.45 & 0.47 & 0.62 & 0.50 & 0.40 & 1.23 & 1.16 & 20.94 & 2.09 \\
& Image Only & 0.19 & 0.12 & 0.20 & 0.14 & 0.13 & 21.57 & 12.49 & 125.03 & 16.20 \\
\rowcolor{shadecolor} 
\cellcolor{white} & Image+Text & 0.78 & 0.68 & 0.92 & 0.83 & 0.69 & 0.61 & 0.72 & 8.53 & 0.62 \\
\hline
\end{tabular}
\end{adjustbox}
\vspace{-5pt}
\end{table*}


\begin{figure}[t!]
  \begin{minipage}[b]{.63\textwidth}
    \centering
    \begin{minipage}{0.49\linewidth}
      \includesvg[width=\linewidth]{accuracy_heatmap.svg}
    \end{minipage}%
    \hfill
    \begin{minipage}{0.49\linewidth}
      \includesvg[width=\linewidth]{f1_score_heatmap.svg}
    \end{minipage}
    \captionof{figure}{\textbf{\textit{Zero-shot generalization:}} Visualization of accuracy (left) and F1-score (right) for zero-shot cross-dataset performance using BLIP-2. Each heatmap illustrates results from fine-tuning on one dataset (y-axis) and evaluating on others.}
    \label{fig:zero_generalise_heatmap}
  \end{minipage}\hfill
  \begin{minipage}[b]{.35\textwidth}
    \centering
    \footnotesize
    \setlength{\tabcolsep}{3pt}
    \begin{adjustbox}{max width=\textwidth}
    \begin{tabular}{l|cccc}
      \toprule
      \textbf{Models} & \textbf{Zero} & \textbf{ICL} & \textbf{CoT} & \textbf{LoRA} \\
      \midrule
      \rowcolor{shadecolor} 
      GPT-4v & 2/1/2 & 1/3/1 & 1/1/2 & - \\
      \rowcolor{shadecolor} 
      GPT-4o & 5/3/1 & 3/2/2 & 3/2/1 & - \\
      JanusPro 7B & 1/2/3 & 2/1/3 & 2/3/3 & - \\ 
      BLIP-2 & 7/5/8 & 7/4/6 & 7/3/7 & 1/1/3 \\ 
      Llava1.5 13B & 4/8/4 & 4/6/7 & 6/6/4 & 4/4/2 \\ 
      Llama v2 7B & 9/7/9 & 9/8/9 & 9/8/9 & 6/6/6 \\ 
      CogVLM & 4/6/5 & 6/5/4 & 8/5/5 & 3/2/1 \\ 
      Qwen VL & 8/9/7 & 8/9/5 & 5/9/6 & 5/3/4 \\ 
      mPlugOWL2 & 3/4/6 & 5/7/8 & 4/7/8 & 2/5/5 \\ 
      \bottomrule
    \end{tabular}
    \end{adjustbox}
    \captionof{table}{\textbf{\textit{Overall ranking:}} Performance in terms of average ranking across datasets (classification/regression/description).}
    \label{table:ranks}
  \end{minipage}
  \vspace{-5pt}
\end{figure}
\noindent
\textbf{Finetuning:} 
Table \ref{table:performance_lora} presents classification results after model adaptation, showing significant performance improvements with fine-tuning. A similar trend is observed for regression tasks, where adaptation reduces prediction error (Figure \ref{fig:teaser}). Detailed results are provided in the Appendix. Overall, BLIP-2 achieves the best performance across both classification and regression tasks, while mPlugOWL2 remains competitive in classification but underperforms in regression.
Table \ref{table:performance_chebi_lora} show performance on molecular description task after finetuning. We observe that CogVLM outperforms other VLMs across all metrics and performs better than recent graph-based LLM.

\noindent
\textbf{Zero-shot generalization:} 
Figure \ref{fig:zero_generalise_heatmap} shows results with zero-shot generalization where we adapted the model on one dataset and evaluated on others. As shown, the performance is better than few-shot (Table \ref{table:performance_ICL}), however the performance is not as good as LoRA where the model was finetuned on the target dataset (Table \ref{table:performance_lora}). From Figure \ref{fig:zero_generalise_heatmap} we observe that after training on HIV-V dataset we get the best zero-shot average accuracy (61\%) and best zero-shot avg F1-score is observed after training on BBBP-V. ClinTox-V and BBBP-V are the most difficult datasets for zero shot generalization with an average accuracy of 39\% and 40\%, respectively.

\subsection{Comparison with existing methods}
We compare our approach with traditional methods such as XGBoost, RF, GenRA \cite{helman2019transitioning}, and Unimol \cite{lu2023highly}, as well as recent LLM-based property prediction models that rely solely on text (ChemLLMBench \cite{guo2023can} and ChemLLM \cite{zhang2024chemllm}). Table \ref{table:performance_ICL} shows that incorporating visual information leads to better performance than LLMs using text alone. Table \ref{table:performance_ICL_reg} compares VLMs with traditional models, where Janus-Pro, GPT-4o, and GPT-4v achieve competitive results on ESOL and LD50, though performance on QM9 and PCQM4M remains lower due to extensive training of traditional models on these datasets. In classification tasks (Table \ref{table:performance_lora}), VLMs consistently outperform traditional approaches, with the largest gains on challenging datasets like BBBP-V.
We also compare with a recent graph-based domain specific LLM approach Molca \cite{liu2023molca}, and observe that our vision based approach provides better or comparable performance across all tasks and datasets (Table \ref{table:performance_ICL_reg} and \ref{table:performance_chebi_lora}).

\begin{figure*}[t!]
\centering
\begin{minipage}{0.28\textwidth}
    \includesvg[width=\linewidth]{scatter_plot_ld50_janus.svg}
\end{minipage}
\begin{minipage}{0.32\textwidth}
    \includegraphics[width=\linewidth]{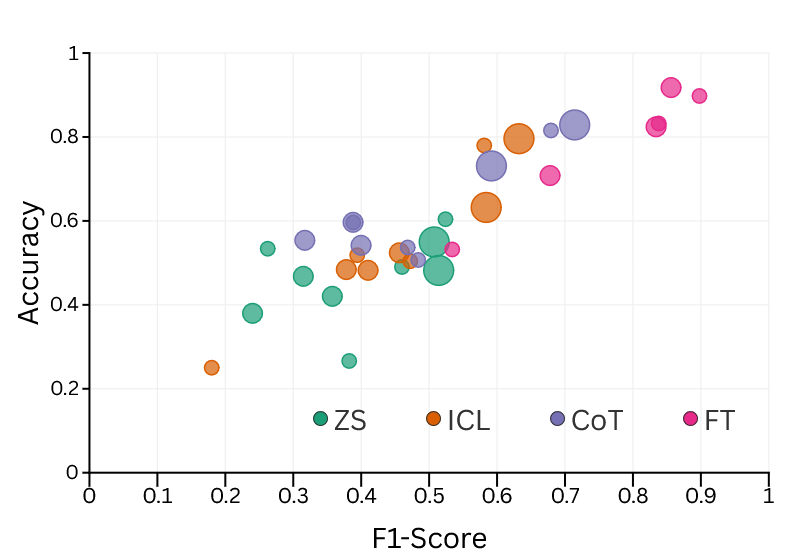}
\end{minipage}
\begin{minipage}{0.37\textwidth}
    \includesvg[width=\linewidth]{line-plot-icl.svg}
\end{minipage}
\vspace{-5pt}
\caption{\textbf{\textit{Analysis on molecular-size, model-size and effect of in-context examples:}} The first plot shows the impact of molecular size on regression error in LD50 with JanusPro, highlighting how visual data improves performance. The middle figure shows comparison of VLMs across datasets  Accuracy vs F1 Score for zero shot (ZS), in-context (ICL), CoT and finetuning (FT).
The bubble size represents the model’s parameter scale. The right figure shows variation in accuracy and f1-scores in case of different ICL examples for GPT-4o model. 
}
\label{fig:reg_ablation_size}
\vspace{-10pt}
\end{figure*}

\begin{figure*}[t!]
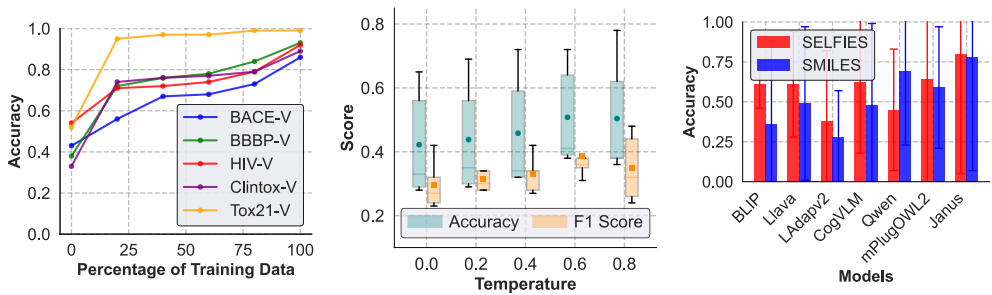

\centering
\begin{minipage}{0.31\textwidth}
    \includesvg[width=\linewidth]{amount_finetune_blip.svg}
\end{minipage}
\begin{minipage}{0.32\textwidth}
    \includesvg[width=\linewidth]{box_plot_temperature.svg}
\end{minipage}
\begin{minipage}{0.31\textwidth}
    \includesvg[width=\linewidth]{smilesvsselfies.svg}
\end{minipage}
\vspace{-8pt}
\caption{\textbf{\textit{Analysis on finetuning, temperature and SELFIES vs. SMILES:}} The first plot shows the impact of percentage of finetuning data. The middle figure shows performance variation with temperature across datasets for BLIP2. The last figure shows analysis of SMILE vs SELFIES string for ICL k=2 across various models.
}
\label{fig:analysis}
\vspace{-15pt}
\end{figure*}



\vspace{-5pt}
\subsection{Discussion and analysis}\label{sec:discussion}
\vspace{-5pt}
This section provides further discussion and analysis for more insights into the benchmark.
\\
\textbf{Impact of visual data:}
Previously, we showed that VLMs outperform LLMs (Tables \ref{table:performance_ICL} and \ref{table:performance_ICL_reg}). Here, in Table \ref{tab:ablation_multi}, we analyze the impact of visual information within the same model by evaluating BLIP-2 and JanusPro in three configurations: (1) Image+Text (molecular structure images + SMILES), (2) Text Only (SMILES), and (3) Image Only (molecular images). We see that image-only inputs are insufficient, but augmenting text with visual data improves performance, showing the benefits of multimodal learning. Limitations and ethical considerations are discussed in Appendix.


\noindent
\textbf{Molecular size:}
We analyze the impact of molecular size on performance and find that larger molecules are more challenging, suggesting that longer representations are harder for models to interpret (Figure \ref{fig:reg_ablation_size} (a)). We also observe that incorporating visual representations improves performance on larger molecules, further reinforcing the value of structural images in property prediction.
\noindent

\textbf{Capability of different models:}
After fine-tuning (Table \ref{table:performance_lora}), BLIP-2 achieves the highest accuracy across classification tasks, except for ClinTox-V, where mPlugOWL2 performs comparably. mPlugOWL2 ranks second overall, followed by CogVLM and Llava 1.5 13B, as confirmed in Table \ref{table:ranks}. BLIP-2 also excels in regression tasks. 



\noindent
\textbf{Impact of model size on performance:}
Since VLMs are trained on large-scale datasets, their size generally correlates with performance. In our analysis of property prediction, we observe a similar trend, where larger models consistently outperform their smaller counterparts (Figure \ref{fig:reg_ablation_size} (c)). Notably, while larger models perform better post-ICL, smaller models surpass them after fine-tuning.

\noindent
\textbf{Effect of number of ICL examples:}
Figure \ref{fig:reg_ablation_size} (a) shows that performance improves with more in-context examples but degrades beyond a certain point. This can be accredited to VLMs' limitations in processing long prompts with excessive tokens.


\noindent
\textbf{Effect of amount of finetuning data:}
Figure \ref{fig:analysis} (left) show the impact of increasing finetuning data from 0\% to 100\% in 20\% increments. We observe best performance with 100\% data in all datasets, resulting in $\sim$40\% increase in performance.

\noindent
\textbf{Effect of temperature:}
Figure \ref{fig:analysis} shows how accuracy (F1-score) averaged across datasets vary with change in temperature. 
With most of the models (in supplementary) usually highest accuracy and F1-score is observed at lower temperatures (0.0-0.4) (in appendix), however, BLIP-2 showed better performance at higher temperatures (0.8).

\noindent
\textbf{SELFIES vs SMILES.}
SELFIES are more robust molecular representations, adhering to valence and ring constraints, thus avoiding invalid molecule generation \cite{krenn2022selfies}. In Figure \ref{fig:analysis} (right) we observe that with few exceptions, SELFIES provides better scores on most datasets. 

\begin{figure*}[t!]
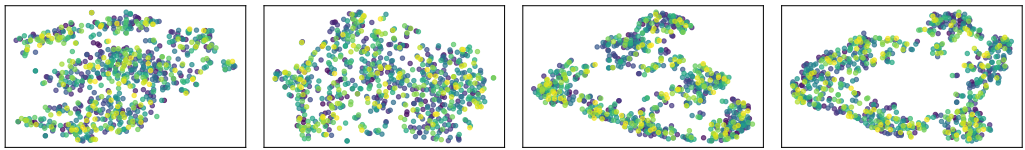

\centering
\begin{minipage}{0.24\textwidth}
    \includesvg[width=\linewidth]{test_vision_encoder_tsne.svg}
\end{minipage}
\begin{minipage}{0.24\textwidth}
    \includesvg[width=\linewidth]{test_multimodal_fusion_tsne.svg}
\end{minipage}
\begin{minipage}{0.24\textwidth}
    \includesvg[width=\linewidth]{test_vision_encoder_tsne_contrastive.svg}
\end{minipage}
\begin{minipage}{0.24\textwidth}
    \includesvg[width=\linewidth]{test_multimodal_fusion_tsne_contrastive.svg}
\end{minipage}
\vspace{-5pt}
\caption{\textbf{\textit{Analyzing visual features:}} The left two plots show t-SNE visualizations of visual encodings of BLIP-2 before and after cross-modal fusion respectively. The right two plots show corresponding t-SNE plots with the proposed contrastive loss using Tanimoto augmentation (T-Aug).
}
\label{fig:tsne_analysis}
\vspace{-5pt}
\end{figure*}




\subsection{Adaptation of vision encoder to molecular structures}
To enhance the visual representation capability of VLMs in the molecular domain, we analyzed the vision embeddings of BLIP-2 using t-SNE and found them to be poorly clustered and non-discriminative—likely due to pretraining on natural images (Fig.~\ref{fig:tsne_analysis}). To address this, we fine-tuned the vision encoder using a contrastive learning objective (NT-Xent loss \cite{chen2020simple}) along with LoRA finetuning. We follow two strategies for identifying the positive pairs and use them in separate approaches. First approach uses augmented views of the same molecule (Aug, Table \ref{table:contrastive_learning_performance}), and second approach uses structurally similar molecules identified via Tanimoto similarity ($>$0.85) (T-Aug). As shown in Fig.~\ref{fig:tsne_analysis}, the similarity-based approach leads to more distinct and meaningful clusters in the embedding space, capturing pharmacophoric patterns more effectively. This method significantly improves performance—reducing ESOL RMSE by 35\%, LD50 MAE by 51\%, and boosting classification accuracy and F1 scores by 2–4\% over the base model. These results underscore the importance of domain-aware vision adaptation, and demonstrate that Tanimoto-guided contrastive learning offers a simple yet powerful enhancement for VLMs in molecular property prediction.

\begin{table}[t!]
    \caption{\textbf{\textit{Performance comparison for proposed contrastive learning:}} Evaluation of different contrastive learning approaches across multiple molecular datasets. 
    }
    \label{table:contrastive_learning_performance}
    \centering \footnotesize
    \begin{adjustbox}{max width=\textwidth,center}
    \begin{tabular}{lcccccccccc}
        \toprule
        \textbf{Method} & \textbf{BACE-V} & \textbf{BBBP-V} & \textbf{HIV-V} & \textbf{Clintox-V} & \textbf{Tox21-V} & \textbf{ESOL-V} & \textbf{LD50-V} & \textbf{QM9-V} & \textbf{PCQM-V} & \textbf{Chebi-V} \\
        & \textbf{Acc(F1)} & \textbf{Acc(F1)} & \textbf{Acc(F1)} & \textbf{Acc(F1)} & \textbf{Acc(F1)} & \textbf{(RMSE)} & \textbf{(MAE)} & \textbf{(MAE)} & & \textbf{(Average)} \\
        \midrule
        LoRA & 0.86(0.83) & 0.93(0.96) & 0.92(0.76) & 0.89(0.93) & 0.99(0.80) & 1.76 & 0.78 & 4.92 & 0.24 & 58.59 \\
        \midrule
        Aug & 0.87(0.85) & 0.94(0.95) & 0.93(0.78) & 0.90(0.94) & 0.97(0.83) & 0.90 & 0.20 & 4.09 & 0.21 & 60.98 \\
        T-Aug & \textbf{0.91(0.88)} & \textbf{0.95(0.96)} & \textbf{0.95(0.84)} & \textbf{0.93(0.93)} & \textbf{0.98(0.89)} & \textbf{0.58} & \textbf{0.10} & \textbf{2.95} & \textbf{0.12} & \textbf{63.73} \\
        \bottomrule
    \end{tabular}
    \end{adjustbox}
    \vspace{-10pt}
\end{table}

\vspace{-5pt}
\section{Conclusion}
\vspace{-5pt}

We present MolVision, a multimodal approach for molecular property prediction using vision-language models. 
We analyze zero-shot, few-shot, and fine-tuned models, combining 2D molecular structure images with textual representations. We provide evaluations on a wide variety of datasets covering classification, regression and description tasks, and demonstrate the benefits of visual information for molecular property prediction.
Adaptation of vision encoder of VLMs to molecular data makes them more promising. This study will serve as a benchmark for further research exploring the use of easily available 2D visual information in multimodal molecular modeling. 

\clearpage

\section{Acknowledgment}
This research has benefitted from the Microsoft Accelerating Foundation Models Research (AFMR) grant program.

\bibliographystyle{ieeetr}
\bibliography{cite}

\clearpage
\appendix
\startcontents[appendix]
\printcontents[appendix]{}{0}{\section*{Table of Contents}}
\clearpage
In this supplementary material, we begin with important discussions on limitations (Section \ref{sec:limitations}) and ethical considerations (Section \ref{ref:ethical}). We have also included additional dataset details (Section \ref{Datasets: Additional Details}), comprehensive results and analysis for classification tasks (Section \ref{Classification}), regression tasks (Section \ref{ref:Regression}), and molecular description tasks (Section \ref{ref:molecular_description}). Furthermore, we present our approach to contrastive learning for vision encoders (Section \ref{ref:contrastive_learning}) and provide detailed prompt examples (Section \ref{prompt_examples}) supporting our discussion in the main paper.

\begin{figure*}[h]
\centerline{\includegraphics[width=0.9\textwidth]{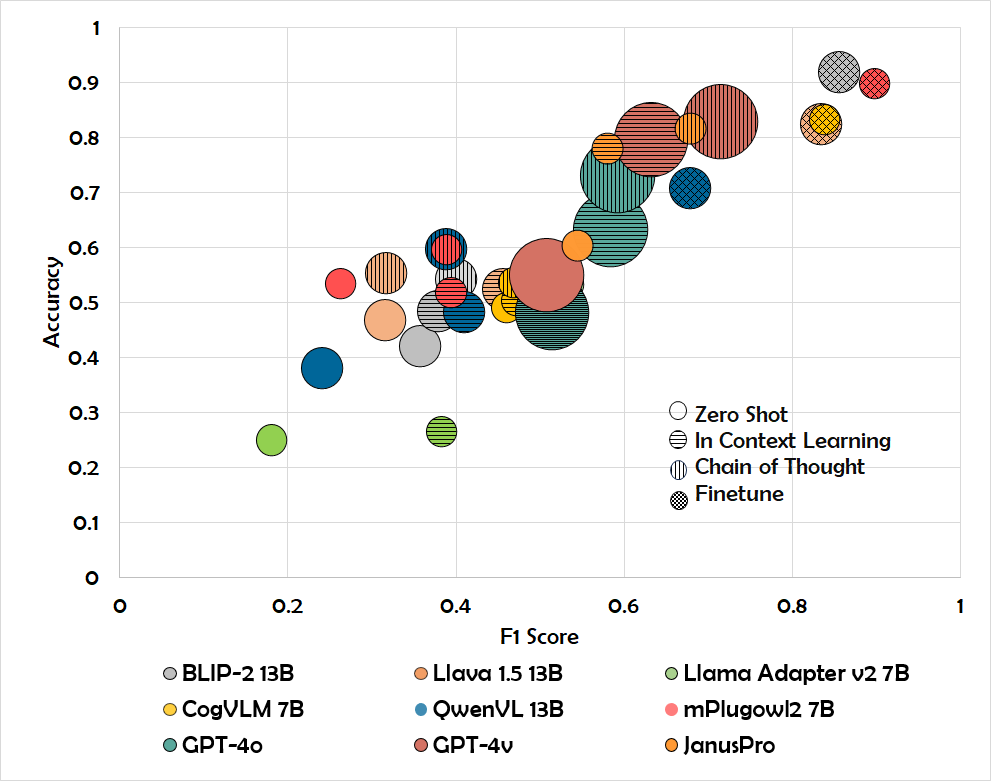}}
\caption{Performance comparison of Vision-Language Models (VLMs) across BACE, BBBP, HIV, Clintox, and Tox21 datasets, depicting Accuracy vs F1 Score. The bubble size represents the model's parameter scale}
\label{fig:bubble_plot_full}
\end{figure*}

\section{Limitations}\label{sec:limitations}
Here we discuss some of the limitations of our work. 

\noindent
\textbf{Adaptation of closed-source models:} Our efficient adaptation of large visual language models for molecular property prediction is limited to open-source models. Considering the strong performance of proprietary models in case of few-shot learning, it will interesting to see how the capabilities of these closed-source models improve for this domain.


\noindent
\textbf{Advanced vision-language models:} In our few-shot setup, we utilize an image representation of a molecule as additional input to the model. Since these models can take only one image as input, it was not possible to provide image representations for in-context examples as input. Future research could explore models capable of processing multiple images simultaneously. 

\section{Ethical considerations}\label{ref:ethical}
The integration of Vision-Language Models (VLMs) into molecular property prediction opens exciting new possibilities while also highlighting the importance of ethical considerations. By leveraging visual and textual representations, these models have the potential to accelerate discoveries in drug development and materials science in a more data-efficient manner. However, ensuring responsible AI development is crucial—focusing on model interpretability, fairness, and transparency can enhance trust and reliability in scientific applications. Additionally, proactive measures, such as open benchmarking and ethical guidelines, can help steer this technology toward positive societal impact while mitigating risks. By addressing these considerations thoughtfully, VLMs can become a transformative tool for chemistry and beyond.

\section{Datasets: Additional Details}
\label{Datasets: Additional Details}
\begin{figure}[h]
\centerline{\includesvg[width=0.5\textwidth]{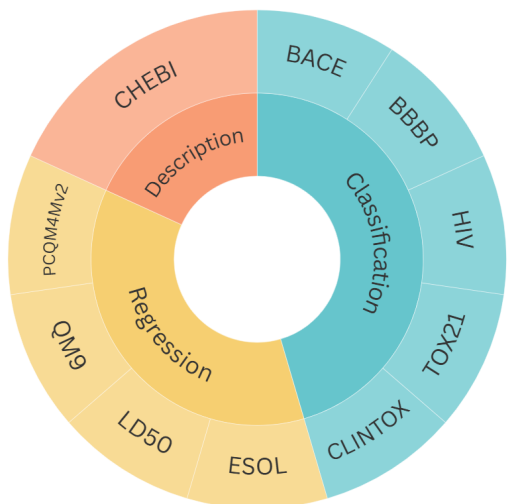}}
\caption{\textbf{\textit{Distribution of Datasets by Task Type}}: The chart illustrates the categorization of datasets based on their primary task, either classification (blue) or regression (yellow).}
\label{fig:dataset_sunburst}
\end{figure}

 This study covers datasets with varied numbers of molecules, as low as 2k to as high as 3.7M. Figure \ref{fig:dataset_sunburst} shows categorization of these datasets by tasks, Classification, Regression and Molecular Description. 
The default representation that is included with these datasets is SMILES and we generated the corresponding SELFIES representation and performed additional evaluations. 
All the data will be available on the provided \href{https://molvision.github.io/MolVision/}{link} \footnote{Code and datasets available at: \url{https://molvision.github.io/MolVision/}}.

\subsection{Model variants}\label{subsec:model_variants}
VLMs are evolved from LLMs that are designed to understand visual information and generate language based on both, textual and visual inputs. They integrate vision encoders and natural language processing techniques to interpret and describe images enabling use cases such as image captioning, visual question answering and multimodal translation. 
We study nine different state-of-the-art visual language models in this study. 


\noindent
\textbf{Janus-Pro 7B:}
Janus-Pro 7B is a unified multimodal model that employs a novel decoupled visual encoding approach. It processes visual inputs differently for understanding versus generation tasks: using a SigLIP encoder to extract semantic features for understanding, while employing a VQ tokenizer for generation. These distinct visual representations are then mapped through separate adaptors into a shared input space, where a 7B-parameter autoregressive transformer processes the combined multimodal sequences. \\
\textbf{BLIP-2:} 
BLIP-2 (Bootstrapping Language-Image Pre-training) is a multimodal model developed by Salesforce that combines visual and language modalities to improve performance on tasks involving both visual inputs and textual information generation.\\
\textbf{Llava 1.5:} 
It is a multimodal model that integrates text and image data, excelling in tasks like Visual Question Answering (VQA), image captioning, and cross-modal retrieval. The model uses Vicuna v1.5 as the base LLM.\\
\textbf{Llama Adapter V2:} The LLaMA-Adapter V2 is an adaption technique that is intended to improve the LLaMA model's ability to obey instructions while preserving parameter efficiency. It presents a number of important methods, such as early fusing of visual knowledge, joint training with discontinuous parameters, bias control of linear layers, and integration with expert models. \\
\textbf{CogVLM:} 
CogVLM is a vision-language model that integrates a Vision Transformer (ViT) encoder, MLP adapter, pretrained large language model, and a visual expert module. The ViT encoder uses the pretrained EVA2-CLIP-E model with the final layer removed for image feature compatibility.\\
\textbf{QwenVL:} 
Qwen-VL is a vision-language model for tasks like understanding, localization, and text reading. It consists of a visual encoder, a position-aware vision-language converter, and a large language model (Qwen-7B). The visual encoder, based on Openclip's ViT-bigG, processes images by dividing them into patches.\\
\textbf{mPlugOwl 2:} 
mPLUGOWL2 integrates a vision encoder, visual abstractor, and language decoder for vision-language tasks. The ViT-L/14 encoder processes images into visual tokens, which the LLaMA-2-7B decoder converts into text. \\
\textbf{GPT-4V:} 
GPT-4V (GPT-4 with Vision) advances multimodal AI by processing both visual and textual inputs. Though its architecture is proprietary, GPT-4V excels in understanding and describing images, solving visual problems, and performing detailed visual-language reasoning across various domains.\\
\textbf{GPT-4o:} 
GPT-4o, OpenAI's advanced language model, improves upon its predecessors with enhanced natural language processing, reasoning, and task completion. It offers better reliability, safety, and zero-shot generalization, though its architecture details remain largely undisclosed.

\subsection{Task and Datasets}\label{subsec:tasks_and_datasets}

We utilize RDKit to generate molecular visualizations from SMILES structures. RDKit not only facilitates the conversion of SMILES strings into visual representations but also supports the transformation of SMILES into SELFIES strings. This functionality enables us to explore diverse molecular encoding techniques, thereby enhancing the robustness and adaptability of our predictive models. Since most existing datasets primarily feature SMILES strings, the ability to convert them to SELFIES representations extends the scope of our analysis. Each dataset contains a formatted prompt alongside the pathway to the visualized molecule image. Leveraging these datasets, we input the information into vision-language models for tasks such as visual question answering and instruction-based challenges. We use the following datasets in our benchmark.



\noindent
\textbf{BACE-V:} 
The BACE-V dataset, adapted from the BACE (Binary Activity of Chemical Entities) dataset, provides 2D skeletal images of molecular structures along with key bioactivity data. Widely used for binary classification in bioactivity prediction, particularly for BACE-1 inhibitors linked to Alzheimer's, the dataset includes both quantitative (IC50 values) and qualitative (binary) binding data. It features 154 BACE inhibitors for affinity prediction, 20 for pose prediction, and 34 for free energy prediction.

\noindent
\textbf{BBBP-V:} 
The BBBP-V dataset, based on the Blood-Brain Barrier Penetration (BBBP) dataset, includes 2D skeletal images of molecular structures. It provides binary labels for BBB penetration (penetrant or non-penetrant) along with SMILES notations and key properties like molecular weight, lipophilicity (logP), and topological polar surface area (TPSA), all essential for predicting BBB permeability.

\noindent
\textbf{HIV-V:} 
The HIV-V dataset, based on the HIV dataset, contains 2D skeletal images of molecular structures to support predictions of HIV replication inhibition. It includes binary labels for anti-HIV activity and key molecular parameters—molecular weight, logP, and TPSA—essential for assessing bioactivity and pharmacokinetics. Our evaluation focused on predicting HIV activity.

\noindent
\textbf{Clintox-V:} 
The ClinTox-V dataset, derived from the ClinTox dataset, includes 2D skeletal images of molecular structures to support predictions of clinical toxicity and FDA approval status. Represented by SMILES notation, each of the 1,491 compounds is labeled for toxicity or FDA approval, enabling two classification tasks. Our evaluation focused on predicting FDA approval status.

\noindent
\textbf{Tox21-V:} 
The Tox21-V dataset, based on the Tox21 dataset, includes 2D skeletal molecular images for predicting chemical toxicity, critical for environmental and pharmaceutical safety. It contains hundreds of compounds, each represented by SMILES notation, with twelve binary labels from toxicological tests. Our evaluation focused on the NR-AR binary label.

\noindent
\textbf{ESOL-V:}
The ESOL-V dataset, based on the ESOL (Estimating Solubility of Organic Compounds in Water) dataset, includes 2D skeletal molecular images and key data for predicting aqueous solubility of organic compounds, crucial for drug development and environmental studies.

\noindent
\textbf{LD50-V:}
It is based on the LD50 (Lethal Dose 50) dataset and includes 2D skeletal molecular images and data on acute toxicity. It focuses on the dose required to cause death in 50\% of test subjects, a key metric for safety assessment in drug development and environmental health.

\noindent
\textbf{QM9-V:}
The QM9-V dataset, derived from the QM9 dataset, contains 2D skeletal representations of molecular structures in image format, alongside various quantum chemical properties. QM9 provides extensive data on 12 quantum mechanical properties, including the dipole moment, isotropic polarizability, electronic spatial extent, HOMO (Highest Occupied Molecular Orbital energy), LUMO (Lowest Unoccupied Molecular Orbital energy), and HOMO-LUMO gap, among others. 

\noindent
\textbf{PCQM4Mv2-V:}
The PCQM4Mv2-V dataset, derived from the PCQM4Mv2 dataset, contains 2D skeletal molecular images paired with quantum property data. The dataset focuses on predicting the HOMO-LUMO gap, an essential quantum property that provides insights into a molecule’s chemical stability and reactivity. 

\textbf{ChEBI-V:}
The ChEBI-V dataset, derived from the ChEBI database, contains 2D skeletal representations of molecular structures in image format, alongside comprehensive biological and chemical annotations. ChEBI-V provides structured information including molecular names, functional classifications, physicochemical properties, and biological roles such as enzyme inhibitors, receptor agonists, and therapeutic agents, making it valuable for molecular description and biological function prediction tasks.

\section{Classification: Further Analysis and Discussion}
\label{Classification}
In this section, we discuss Zero-shot evaluation, effect of number of examples used in ICL,chain of thought prompting, effect of temperature in model performance and impact of visual data for classification task. Figure \ref{fig:bubble_plot_full} show performance comparison of different models.

\subsection{Zero-Shot Evaluation}\label{ref:zeroshot_classification}

We have included more detailed results with Zero-shot performance (Table \ref{table:zeroshot_smile_model_performance} and \ref{table:zero_shot_prompting}) where we only ask questions with general outline to the models without using any in-context examples.

\begin{table*}[!htbp]
    \caption{\textbf{\textit{Zero-shot with SMILES:}} The table shows variation in the F1-score \& accuracy of different models when subjected to zero-shot prompting. In this evaluation, only the basic instruction is provided to the models to predict whether a given molecule string is toxic or not, without any additional context or examples. }
    \label{table:zeroshot_smile_model_performance}
    \centering \small
    \begin{adjustbox}{max width=\textwidth}
    \begin{tabular}{lccccc}
        \toprule
        \textbf{Model} & \textbf{BACE-V} & \textbf{BBBP-V} & \textbf{HIV-V} & \textbf{Clintox-V} & \textbf{Tox21-V} \\ 
        \midrule 
        JanusPro 7B & 0.45(0.37) & 0.42(0.44) & 0.52(0.32) & 0.31(0.32) & 0.44(0.34) \\
        BLIP2 & 0.28 (0.29) & 0.31 (0.29) & 0.42 (0.33) & 0.29 (0.28) & 0.42 (0.31) \\ 
        Llava 1.5 & 0.37 (0.54) & 0.43 (0.46) & 0.38 (0.35) & 0.36 (0.39) & 0.46 (0.39) \\ 
        Llama & 0.34 (0.39) & 0.41 (0.28) & 0.21 (0.33) & 0.28 (0.31) & 0.12 (0.13) \\ 
        CogVLM & 0.27 (0.34) & 0.31 (0.32) & 0.22 (0.25) & 0.47 (0.49) & 0.17 (0.12) \\ 
        QwenVLM & 0.32 (0.39) & 0.29 (0.12) & 0.22 (0.29) & 0.22 (0.15) & 0.45 (0.37) \\ 
        mPlugowl2 & 0.39 (0.38) & 0.32 (0.31) & 0.41 (0.27) & 0.27 (0.26) & 0.67 (0.13) \\ 
        \bottomrule
    \end{tabular}
    \end{adjustbox}
\end{table*}

\begin{table*}[!htbp]
    \caption{\textbf{\textit{Zero-shot performance with SELFIES:}} The table illustrates the variation in the Accuracy (F1-score) of different models when subjected to zero-shot prompting. In this evaluation, only the basic instruction is provided to the models to predict whether a given molecule string is toxic or not, without any additional context or examples. }
    \label{table:zero_shot_prompting}
    \centering \small
    \begin{adjustbox}{max width=\textwidth}
    \begin{tabular}{lccccc}
        \toprule
        \textbf{Model} & \textbf{BACE-V} & \textbf{BBBP-V} & \textbf{HIV-V} & \textbf{ClinTox-V} & \textbf{Tox21-V} \\ 
        \midrule
        JanusPro 7B &0.48(0.53) & 0.54(0.581) & 0.43 (0.36) & 0.48 (0.37) & 0.47 (0.31) \\ 
        BLIP2 & 0.41 (0.34) & 0.46 (0.48) & 0.31 (0.29) & 0.45 (0.47) & 0.57 (0.21) \\ 
        Llava 1.5 & 0.42 (0.48) & 0.48 (0.50) & 0.24 (0.33) & 0.54 (0.64) & 0.59 (0.15) \\ 
        Llama Adapter v2 7B & 0.42 (0.51) & 0.39 (0.42) & 0.21 (0.33) & 0.41 (0.53) & 0.14 (0.12) \\ 
        CogVLM & 0.42 (0.49) & 0.49 (0.62) & 0.28 (0.39) & 0.44 (0.35)  & 0.16 (0.11) \\ 
        QwenVL & 0.42 (0.59) & 0.41 (0.58) & 0.21 (0.31) & 0.16 (0.14)  & 0.16 (0.11) \\ 
        mPlugOwl2 & 0.47 (0.25) & 0.35 (0.23) & 0.39 (0.28) & 0.28 (0.22) & 0.37 (0.17) \\ 
        \bottomrule
    \end{tabular}
    \end{adjustbox}
\end{table*}

\textbf{SMILES vs SELFIES:} We examine and compare Zero-shot performance of models with SMILES and SELFIES representations. SELFIES generally yield better performance however on HIV dataset we see comparatively better performance with SMILES representation  
as shown in Tables~\ref{table:zeroshot_smile_model_performance} and~\ref{table:zero_shot_prompting}. We also performed this analysis with ICL and has been discussed later.

\subsection{Effect of ICL Examples}\label{ref:few_shot_classification}
We conducted a comprehensive analysis of effects of number of examples (k = 0, 2, 4) in in-context learning (ICL) across vision-language models for molecular property prediction. More context does not always yield better results (Table \ref{table:icl_k0}, \ref{table:icl_k4}). The effectiveness of ICL varies significantly across datasets, as evidenced by CogVLM's substantial improvement on ClinTox-V when increasing from k = 0 to k = 4 (0.54 to 0.76 accuracy). We also observed similar behavious with CogVLM using SELFIES in Table~\ref{table:effect_in_context_examples}. Different models demonstrate varying sensitivity to ICL,  BLIP-2 however show consistent improvement with increased context, achieving its best performance with k = 4 across most datasets (Table~\ref{table:blip2_icl_variation}).   

Model QwenVL shows peak performance with k = 2 on several datasets (Table~\ref{table:effect_in_context_examples_QwenVL}). 
Comparing SMILES representations (Table \ref{table:icl_k0}, \ref{table:icl_k4} ) versus SELFIES representations (Table \ref{table:effect_in_context_examples_llava}, \ref{table:effect_in_context_examples_mPlugOwl2},\ref{table:effect_in_context_examples_QwenVL} \ref{table:effect_in_context_examples_Llava_Adapter_V2_7B}, \ref{table:effect_in_context_examples_BLIP2}), SELFIES maintains more stable performance across different k values, particularly for complex models like mPlugOwl2 and CogVLM. These findings indicate that ICL's effectiveness depends heavily on model architecture, molecular representation, and dataset characteristics.

We also included results with increased in-context examples with gpt-4o. With the exception of BBBP-V accuracy improved across all datasets with an increase in the number of in-context examples (k) to six or eight. Notably, on the BACE-V, and Clintox-V datasets, we observed approximately a 40\% increase in accuracy. With the exception of BACE-V, the F1-score was also highest at k=2 or 4 across all datasets (Table \ref{table:numberICLexamples}).

\begin{table*}[!htbp]
    \caption{\textbf{\textit{Role of in-context examples:}} ICL with k = 0 showing Accuracy (F1-score) of various models on different datasets with SMILES representations.}
    \label{table:icl_k0}
    \centering \small
    \begin{adjustbox}{max width=\textwidth}
    \begin{tabular}{lccccc}
        \toprule
        \textbf{Model} & \textbf{BACE-V} & \textbf{BBBP-V} & \textbf{HIV-V} & \textbf{ClinTox-V} & \textbf{Tox21-V} \\ 
        \midrule
        JanusPro 7B & 0.65(0.68) & 0.52(0.51) & 0.92(0.68) & 0.41(0.31) & 0.52(0.44) \\
        BLIP2 & 0.36 (0.52) & 0.33 (0.27) & 0.41 (0.36) & 0.37 (0.28) & 0.63 (0.29) \\ 
        Llava 1.5 & 0.55 (0.18) & 0.47 (0.43) & 0.35 (0.32) & 0.33 (0.33) & 0.64 (0.11) \\ 
        Llama & 0.39 (0.59) & 0.36 (0.42) & 0.21 (0.33) & 0.22 (0.19) & 0.15 (0.19) \\ 
        CogVLM & 0.39 (0.56) & 0.48 (0.48) & 0.39 (0.26) & 0.54 (0.54) & 0.65 (0.11) \\ 
        QwenVLM & 0.41 (0.52) & 0.31 (0.11) & 0.28 (0.21) & 0.38 (0.12) & 0.52 (0.43) \\ 
        mPlugowl2 & 0.48 (0.16) & 0.41 (0.43) & 0.65 (0.18) & 0.28 (0.28) & 0.85 (0.11) \\ 
        \bottomrule
    \end{tabular}
    \end{adjustbox}
\end{table*}

\begin{table*}[!htbp]
    \caption{\textbf{\textit{Effect of in-context examples:}} ICL with k = 4 showing Accuracy (F1-score) of various models on different datasets with SMILES Representation.}
    \label{table:icl_k4}
    \centering \small
    \begin{adjustbox}{max width=\textwidth}
    \begin{tabular}{lccccc}
        \toprule
        \textbf{Model} & \textbf{BACE-V} & \textbf{BBBP-V} & \textbf{HIV-V} & \textbf{ClinTox-V} & \textbf{Tox21-V} \\ 
        \midrule
        JanusPro 7B & 0.73(0.62) & 0.63(0.601)) & 0.95(0.69) & 0.97(0.64) & 0.71(0.62) \\
        BLIP2 & 0.29 (0.42) & 0.19 (0.11) & 0.52 (0.32) & 0.34 (0.36) & 0.55 (0.39) \\ 
        Llava 1.5 & 0.48 (0.38) & 0.57 (0.66) & 0.32 (0.33) & 0.25 (0.22) & 0.64 (0.25) \\ 
        Llama & 0.39 (0.56) & 0.37 (0.22) & 0.21 (0.33) & 0.32 (0.19) & 0.28 (0.12) \\ 
        CogVLM & 0.39 (0.54) & 0.64 (0.34) & 0.39 (0.26) & 0.68 (0.48) & 0.19 (0.11) \\ 
        QwenVLM & 0.42 (0.52) & 0.31 (0.11) & 0.42 (0.54) & 0.81 (0.12) & 0.72 (0.17) \\ 
        mPlugowl2 & 0.58 (0.42) & 0.43 (0.38) & 0.71 (0.25) & 0.38 (0.42) & 0.83 (0.13) \\ 
        \bottomrule
    \end{tabular}
    \end{adjustbox}
\end{table*}

\begin{table*}[!htbp]
    \caption{\textbf{\textit{Effect of in-context examples:}} Accuracy (F1-score) of BLIP-2 Model using SELFIES representations with variation in number of in-context examples used in the prompt (k = 0, 2, 4).
    }
    \label{table:blip2_icl_variation}
    \centering \small
    \begin{adjustbox}{}
    \begin{tabular}{lccccc}
        \toprule
        \textbf{Variation} & \textbf{BACE-V} & \textbf{BBBP-V} & \textbf{HIV-V} & \textbf{ClinTox-V} & \textbf{Tox21-V} \\ 
        \midrule
        k = 0 & 0.43(0.26) & 0.38(0.27) & 0.54(0.32) & 0.33(0.42) & 0.52(0.21) \\ 
        k = 2 & 0.36\textbf{(0.52)} & 0.37(0.29) & 0.60(0.29) & 0.34(0.36) & 0.75(0.42) \\ 
        k = 4 & \textbf{0.61}(0.27) & \textbf{0.39(0.31)} & \textbf{0.81(0.39)} & \textbf{0.36(0.44)} & \textbf{0.79(0.48)} \\ 
        \bottomrule
    \end{tabular}
    \end{adjustbox}
\end{table*}

\begin{table*}[!htbp]
    \caption{\textbf{\textit{Impact of number of in-context examples:}} The table illustrates the variation in the performance of the CogVLM model with ICL in terms of Accuracy (F1-score), which utilizes the Vicuna 7B as its backbone, when tested on the SELFIE representation of various datasets. The performance is evaluated with different numbers of in-context examples (k = 0, 2, 4) provided in the prompt. The following results are produced with temperature set to 0.}
    \label{table:effect_in_context_examples}
    \centering \small
    \begin{adjustbox}{max width=\textwidth}
    \begin{tabular}{lccccc}
        \toprule
        \textbf{Variation} & \textbf{BACE-V} & \textbf{BBBP-V} & \textbf{HIV-V} & \textbf{ClinTox-V} & \textbf{Tox21-V} \\ 
        \midrule
        k = 0 & 0.34 (0.51) & 0.36 (0.32) & 0.24 (0.33) & 0.54 (0.65) & 0.26 (0.15) \\ 
        k = 2 & 0.62 (0.44) & 0.51 (0.58) & 0.25 (0.29) & 0.32 (0.37) & 0.18 (0.14) \\ 
        k = 4 & 0.44 (0.53) & 0.38 (0.45) & 0.32 (0.31) & 0.76 (0.86) & 0.47 (0.13) \\ 
        \bottomrule
    \end{tabular}
    \end{adjustbox}
\end{table*}

\begin{table*}[!htbp]
    \caption{\textbf{\textit{Impact of number of in-context examples:}} The table illustrates the variation in the performance of the Llava 1.5 model with ICL in terms of Accuracy (F1-score) using Llava 1.5 13 billion parameters, when tested on the SELFIE representation of various datasets. The performance is evaluated with different numbers of in-context examples (k = 0, 2, 4) provided in the prompt. (Temperature=0).}
    \label{table:effect_in_context_examples_llava}
    \centering \small
    \begin{adjustbox}{max width=\textwidth}
    \begin{tabular}{lccccc}
        \toprule
        \textbf{Variation} & \textbf{BACE-V} & \textbf{BBBP-V} & \textbf{HIV-V} & \textbf{ClinTox-V} & \textbf{Tox21-V} \\ 
        \midrule
        k = 0 & 0.62 (0.24) & 0.35 (0.17) & 0.51 (0.13) & 0.20 (0.14) & 0.76 (0.17) \\ 
        k = 2 & 0.61 (0.33) & 0.73 (0.46) & 0.32 (0.35) & 0.35 (0.36) & 0.67 (0.47) \\ 
        k = 4 & 0.49 (0.38) & 0.56 (0.29) & 0.42 (0.33) & 0.25 (0.19) & 0.89 (0.11) \\ 
        \bottomrule
    \end{tabular}
    \end{adjustbox}
\end{table*}

\begin{table*}[!htbp]
    \caption{\textbf{\textit{Impact of number of in-context examples:}} The table illustrates the variation in the performance of the mPlugOwl2 model with ICL in terms of Accuracy (F1-score). This model utilizes Llama2 7B as its backbone and is tested on the SELFIE representation of various datasets. The performance is evaluated with different numbers of in-context examples (k = 0, 2, 4) provided in the prompt. (Temperature = 0).}
    \label{table:effect_in_context_examples_mPlugOwl2}
    \centering \small
    \begin{adjustbox}{max width=\textwidth}
    \begin{tabular}{lccccc}
        \toprule
        \textbf{Variation} & \textbf{BACE-V} & \textbf{BBBP-V} & \textbf{HIV-V} & \textbf{ClinTox-V} & \textbf{Tox21-V} \\ 
        \midrule
        k = 0 & 0.53 (0.22) & 0.48 (0.52) & 0.57 (0.19) & 0.22 (0.15) & 0.62 (0.17) \\ 
        k = 2 & 0.64 (0.65) & 0.46 (0.46) & 0.76 (0.74) & 0.39 (0.43) & 0.74 (0.21) \\ 
        k = 4 & 0.61 (0.31) & 0.35 (0.36) & 0.73 (0.41) & 0.46 (0.57) & 0.76 (0.14) \\ 
        \bottomrule
    \end{tabular}
    \end{adjustbox}
\end{table*}

\begin{table*}[!htbp]
    \caption{\textbf{\textit{Impact of number of in-context examples:}} The table illustrates the variation in the performance of the QwenVL model with ICL in terms of Accuracy (F1-score) using QwenVL 7 B parameters, when tested on the SELFIE representation of various datasets. The performance is evaluated with different numbers of in-context examples (k = 0, 2, 4) provided in the prompt.(Temperature=0).}
    \label{table:effect_in_context_examples_QwenVL}
    \centering \small
    \begin{adjustbox}{}
    \begin{tabular}{lccccc}
        \toprule
        \textbf{Variation} & \textbf{BACE-V} & \textbf{BBBP-V} & \textbf{HIV-V} & \textbf{ClinTox-V} & \textbf{Tox21-V} \\ 
        \midrule
        k = 0 & 0.41 (0.52) & 0.31 (0.10) & 0.80 (0.11) & 0.18 (0.12) & 0.52 (0.14) \\ 
        k = 2 & 0.45 (0.38) & 0.63 (0.49) & 0.79 (0.49) & 0.50 (0.48) & 0.78 (0.76) \\ 
        k = 4 & 0.42 (0.51) & 0.29 (0.09) & 0.81 (0.10) & 0.42 (0.54) & 0.72 (0.17) \\ 
        \bottomrule
    \end{tabular}
    \end{adjustbox}
\end{table*}

\begin{table*}[!htbp]
    \caption{\textbf{\textit{Impact of number of in-context examples}} The table illustrates the variation in the Accuracy (F1-score) of the ICL model using Llama Adapter V2, which utilizes Llama2 7B as its backbone, when tested on the SELFIE representation of various datasets. The performance is evaluated with different numbers of in-context examples (k = 0, 2, 4) provided in the prompt.(Temperature=0).}
    \label{table:effect_in_context_examples_Llava_Adapter_V2_7B}
    \centering \small
    \begin{adjustbox}{max width=\textwidth}
    \begin{tabular}{lccccc}
        \toprule
        \textbf{Variation} & \textbf{BACE-V} & \textbf{BBBP-V} & \textbf{HIV-V} & \textbf{ClinTox-V} & \textbf{Tox21-V} \\ 
        \midrule
        k = 0 & 0.37 (0.34) & 0.49 (0.21) & 0.21 (0.33) & 0.30 (0.39) & 0.14 (0.17) \\ 
        k = 2 & 0.38 (0.44) & 0.48 (0.23) & 0.21 (0.37) & 0.31 (0.35) & 0.15 (0.17) \\ 
        k = 4 & 0.36 (0.35) & 0.51 (0.27) & 0.31 (0.42) & 0.50 (0.31) & 0.15 (0.18) \\ 
        \bottomrule
    \end{tabular}
    \end{adjustbox}
\end{table*}

\begin{table*}[!htbp]
    \caption{\textbf{\textit{Impact of number of in-context examples:}} The table illustrates the variation in the Accuracy (F1-score) of the ICL model using the BLIP-2 model when tested on the SELFIE representation of various datasets. The performance is evaluated with different numbers of in-context examples (k = 0, 2, 4) provided in the prompt.(Temperature=0).}
    \label{table:effect_in_context_examples_BLIP2}
    \centering \small
    \begin{adjustbox}{max width=\textwidth}
    \begin{tabular}{lccccc}
        \toprule
        \textbf{Variation} & \textbf{BACE-V} & \textbf{BBBP-V} & \textbf{HIV-V} & \textbf{Clintox-V} & \textbf{Tox21-V} \\ 
        \midrule
        k = 0 & 0.56 (0.10) & 0.31 (0.12) & 0.49 (0.24) & 0.16 (0.09) & 0.47 (0.32) \\ 
        k = 2 & 0.61 (0.15) & 0.35 (0.16) & 0.56 (0.24) & 0.18 (0.12) & 0.41 (0.22) \\ 
        k = 4 & 0.66 (0.26) & 0.32 (0.11) & 0.51 (0.29) & 0.21 (0.10) & 0.45 (0.19) \\ 
        \bottomrule
    \end{tabular}
    \end{adjustbox}
\end{table*}

\begin{table*}[!htbp]
    \caption{\textbf{Effect of in-context examples:} Accuracy (F1-score) for different ICL examples on GPT-4o model.
    }
    \label{table:numberICLexamples}
    \centering \small
    \begin{tabular}{lccccc|c}
         \toprule
        \textbf{ICL Variation} & \textbf{BACE-V} & \textbf{BBBP-V} & \textbf{HIV-V} & \textbf{Clintox-V} & \textbf{Tox21-V} & \textbf{Average} \\
        \midrule
         k=0   & 0.39 (\textbf{0.55}) & 0.56 (0.64)   & 0.72 (0.53)   & 0.25 (0.33)   & 0.49 (0.46)   & 0.48/0.50 \\ 
         k=2   & 0.56 (0.53)   & \textbf{0.77} (\textbf{0.81})   & 0.82 (0.56)   & 0.59 (0.44)   & 0.42 (\textbf{0.58})   & 0.63/0.58 \\
         k=4   & 0.64 (0.52)   & 0.63 (0.66)   & 0.82 (0.78)   & 0.71 (\textbf{0.69})   & 0.52 (0.44)   & 0.66/\textbf{0.61} \\ 
         k=6   & 0.61 (0.48)   & 0.56 (0.62)   & \textbf{0.86} (\textbf{0.79})   & \textbf{0.76} (0.63)   & \textbf{0.61} (0.43)   & \textbf{0.68}/0.59 \\
         k=8   & \textbf{0.72} (0.51)   & 0.56 (0.60)   & 0.72 (0.64)   & 0.67 (\textbf{0.69})   & 0.55 (0.34)   & 0.64/0.55 \\ 
         k=10  & 0.55 (0.35)   & 0.55 (0.59)   & 0.69 (0.23)   & 0.49 (0.53)   & \textbf{0.61} (0.27)   & 0.57/0.39 \\ 
        \bottomrule
    \end{tabular}
\end{table*}

\subsection{Chain of Thought Prompting}\label{ref:cot_classification}

Table \ref{table:cot_classification} demonstrates the effectiveness of Chain of Thought (CoT) prompting on molecular property classification tasks across five benchmark datasets. GPT-4v emerges as the top performer with an average accuracy of 72.32\%, closely followed by GPT-4o at 71.14\%, indicating that both commercial models excel when employing structured reasoning approaches. Janus achieves the third-highest performance with an average accuracy of 71.60\%, demonstrating competitive capabilities among open-source models and particularly excelling on HIV-V (93.3\%) and ClinTox-V (97.2\%) datasets. 

The results reveal significant performance variations across datasets, with HIV-V generally showing the highest accuracy scores across models, while BBBP-V and Tox21-V present greater challenges. Notably, QwenVLM shows strong performance on HIV-V (82.9\%) and Tox21-V (73.9\%), while mPlugOWL2 demonstrates exceptional performance on specific datasets like Tox21-V (76.0\%) and HIV-V (75.2\%). The remaining models exhibit moderate performance, with CogVLM achieving balanced results across datasets and BLIP-2 showing consistent but lower performance. 
\begin{table*}[!htbp]
    \caption{\textbf{\textit{Classification Performance in Chain of Thought Prompting:}} Comparisons of models evaluated on classification tasks across various datasets using Chain-of-Thought (CoT) prompting showing Accuracy (F1-score) with SMILES representations.}
    \label{table:cot_classification}
    \centering \small
    \begin{adjustbox}{max width=\textwidth}
    \begin{tabular}{lccccc}
        \toprule
        \textbf{Model} & \textbf{BACE-V} & \textbf{BBBP-V} & \textbf{HIV-V} & \textbf{ClinTox-V} & \textbf{Tox21-V} \\ 
        \midrule
        GPT-4o & 0.783(0.612) & 0.696(0.481) & 0.893(0.829) & 0.683(0.582) & 0.601(0.455) \\
        GPT-4v & 0.800(0.749) & 0.716(0.620) & 0.928(0.842) & 0.972(0.729) & 0.728(0.632) \\
        BLIP-2 & 0.49 (0.52) & 0.49 (0.41) & 0.62 (0.32) & 0.54 (0.36) & 0.57 (0.39) \\ 
        CogVLM & 0.510(0.559) & 0.673(0.396) & 0.422(0.384) & 0.650(0.701) & 0.430(0.303) \\ 
        mPlugOWL2 & 0.716(0.414) & 0.555(0.362) & 0.752 (0.413) & 0.461(0.574) & 0.76 (0.148) \\ 
        Llava & 0.523 (0.462) & 0.582(0.493) & 0.42 (0.33) & 0.352(0.19) & 0.893(0.11) \\ 
        Llama-Adapter & 0.430(0.339) & 0.554(0.674) & 0.429(0.312) & 0.684(0.712) & 0.437(0.382) \\ 
        QwenVLM & 0.528(0.429) & 0.394(0.291) & 0.829(0.329) & 0.492(0.421) & 0.739(0.471) \\ 
        Janus & 0.797(0.669) & 0.696(0.661) & 0.933(0.781) & 0.972(0.729) & 0.681(0.557) \\
        \bottomrule
    \end{tabular}
    \end{adjustbox}
\end{table*}
\subsection{Effect of Temperature}\label{ref:temperature_effect_classification}
The effect of temperature variation was shown in the main paper on one model (BLIP2 model) and here we include more results examining the effect of temperature variation under different settings (Table \ref{table:temperature_effect_smiles}, \ref{table:accuracytemp} Table \ref{table:temperature_variations_cogvlm}, and Table \ref{table:temperature_effect_llava_model}, \ref{table:temperature_effect_selfie}, \ref{table:accuracytemp_blip}). 
We analyzed the impact of sampling temperature (ranging from 0.0 to 0.8) on model performance across different molecular representations and architectures. For SELFIES representation, the Llama Adapter v2 model shows optimal performance at moderate temperatures (0.2-0.4) for the BBBP-V dataset, achieving accuracy of 0.66 at temperature 0.2 (Table~\ref{table:temperature_effect_selfie}). SMILES representation exhibits different temperature sensitivity, with generally improved performance at higher temperatures across datasets (Table~\ref{table:temperature_effect_smiles}). BLIP2 demonstrates consistent improvement with increasing temperature, achieving peak average performance of 0.51 at temperature 0.6 (Table~\ref{table:accuracytemp}). Llava 1.5 13B shows optimal performance at lower temperatures, particularly for the Tox21-V dataset with 0.93 accuracy at temperature 0.2 (Table~\ref{table:temperature_effect_llava_model}). CogVLM exhibits more stable performance across temperature variations, with slight degradation at higher temperatures (Table~\ref{table:temperature_variations_cogvlm}). The mPlugOWL2 model achieves its best performance at temperature 0.2 across multiple datasets, notably reaching 0.88 accuracy on Tox21-V (Table~\ref{table:accuracytemp}). These findings suggest that moderate temperatures (0.2-0.4) generally provide optimal performance across models and representations, with specific optimal values being model and dataset dependent.
\begin{table*}[!htbp]
    \caption{\textbf{Effect of temperature:} Accuracy (F1-score) for different temperature settings using ICL (Samples k=2) on mPlugOWL2 model with SMILES. 
    }
    \label{table:accuracytemp}
    \centering \small
    \begin{adjustbox}{}
    \begin{tabular}{lccccc}
         \toprule
        \textbf{Temp Variation} & \textbf{BACE-V} & \textbf{BBBP-V} & \textbf{HIV-V} & \textbf{Clintox-V} & \textbf{Tox21-V} \\
        \midrule
        0.0 & 0.59(\textbf{0.32}) & 0.35(0.38) & 0.62(\textbf{0.30}) & 0.34(0.42) & 0.69(\textbf{0.57}) \\
        0.2 & \textbf{0.70}(0.28) & 0.38(0.24) & \textbf{0.74}(0.18) & \textbf{0.78}(0.18) & \textbf{0.88}(0.10) \\
        0.4 & 0.65(0.15) & \textbf{0.46(0.46)} & 0.64(0.21) & 0.30(0.36) & 0.83(0.16) \\
        0.6 & 0.60(0.23) & 0.43(0.44) & 0.62(0.26) & 0.40(0.51) & 0.73(0.16) \\
        0.8 & 0.58(0.19) & 0.41(0.43) & 0.59(0.28) & {0.43(\textbf{0.52})} & 0.72(0.20) \\
        \bottomrule
    \end{tabular}
    \end{adjustbox}
\end{table*}
\begin{table*}[!htbp]
    \caption{\textbf{\textit{Effect of temperature on SELFIE:}} This table shows the performance of the Llama Adapter v2 model with ICL (k=2 examples) on various datasets represented in SELFIE notation.}
    \label{table:temperature_effect_selfie}
    \centering \small
    \begin{tabular}{lccccc}
        \toprule
        \textbf{Temp} & \textbf{BACE-V} & \textbf{BBBP-V} & \textbf{HIV-V} & \textbf{Clintox-V} & \textbf{Tox21-V} \\ 
        \midrule
        0.0 & 0.38 (0.44) & 0.48 (0.23) & 0.21 (0.37) & 0.31 (0.35) & 0.15 (0.17) \\ 
        0.2 & 0.34 (0.51) & 0.66 (0.79) & 0.19 (0.31) & 0.29 (0.88) & 0.13 (0.18) \\ 
        0.4 & 0.35 (0.49) & 0.62 (0.75) & 0.22 (0.32) & 0.24 (0.85) & 0.28 (0.11) \\ 
        0.6 & 0.36 (0.43) & 0.53 (0.68) & 0.26 (0.33) & 0.22 (0.83) & 0.31 (0.18) \\ 
        0.8 & 0.33 (0.44) & 0.58 (0.69) & 0.24 (0.39) & 0.22 (0.81) & 0.31 (0.15) \\ 
        \bottomrule
    \end{tabular}
\end{table*}

\begin{table*}[!htbp]
    \caption{\textbf{\textit{Effect of temperature on SMILES:}} This table shows the performance of the Llama Adapter v2 model with ICL (k=2 examples) on various datasets represented in SMILES notation.}
    \label{table:temperature_effect_smiles}
    \centering \small
    \begin{tabular}{lccccc}
        \toprule
        \textbf{Temp} & \textbf{BACE-V} & \textbf{BBBP-V} & \textbf{HIV-V} & \textbf{Clintox-V} & \textbf{Tox21-V} \\ 
        \midrule
        0.0 & 0.28 (0.29) & 0.18 (0.11) & 0.19 (0.17) & 0.29 (0.12) & 0.31 (0.21) \\ 
        0.2 & 0.38 (0.55) & 0.22 (0.83) & 0.21 (0.33) & 0.28 (0.19) & 0.32 (0.15) \\ 
        0.4 & 0.35 (0.49) & 0.35 (0.76) & 0.42 (0.35) & 0.22 (0.24) & 0.39 (0.24) \\ 
        0.6 & 0.39 (0.56) & 0.23 (0.74) & 0.31 (0.25) & 0.31 (0.11) & 0.41 (0.29) \\ 
        0.8 & 0.43 (0.54) & 0.29 (0.62) & 0.37 (0.36) & 0.26 (0.19) & 0.41 (0.11) \\ 
        \bottomrule
    \end{tabular}
\end{table*}
\begin{table*}[!htbp]
    \caption{\textbf{Effect of Temperature on Model Performance:} Accuracy (F1-score) at different temperature settings using the BLIP2 model on various datasets. Higher temperatures generally show increased variability in F1-scores, with peak performance occurring at different temperature levels across datasets.}
    \label{table:accuracytemp_blip}
    \centering \small
    \begin{adjustbox}{}
    \begin{tabular}{lcccccc}
         \toprule
        \textbf{Temp Variation} & \textbf{BACE-V} & \textbf{BBBP-V} & \textbf{HIV-V} & \textbf{Clintox-V} & \textbf{Tox21-V} & \textbf{Average} \\
        \midrule
        0.0 & 0.33(0.42) & 0.29(0.27) & 0.56(0.24) & 0.28(0.23) & 0.65(0.32) & 0.42(0.30) \\
        0.2 & 0.35(0.49) & 0.30(0.28) & 0.56(0.30) & 0.29(0.16) & 0.69(0.34) & 0.44(0.31) \\
        0.4 & 0.34(0.42) & 0.32(0.28) & 0.59(0.27) & 0.32(0.34) & 0.72(0.34) & 0.46(0.33) \\
        0.6 & 0.41(0.54) & 0.39(0.31) & 0.64(0.35) & 0.38(0.38) & 0.72(0.36) & 0.51(0.39) \\
        0.8 & 0.38(0.48) & 0.36(0.24) & 0.62(0.32) & 0.38(0.26) & 0.78(0.44) & 0.50(0.35) \\
        \bottomrule
    \end{tabular}
    \end{adjustbox}
\end{table*}

\begin{table*}[!htbp]
    \caption{\textbf{\textit{Effect of temperature:}} Performance analysis using the Llava 1.5 13B with ICL (k=2), focusing on diverse datasets represented in SELFIE format.}
    \label{table:temperature_effect_llava_model}
    \centering \small
    \begin{adjustbox}{}
    \begin{tabular}{lccccc}
        \toprule
        \textbf{Temp} & \textbf{BACE-V} & \textbf{BBBP-V} & \textbf{HIV-V} & \textbf{Clintox-V} & \textbf{Tox21-V} \\ 
        \midrule
        0.0 & 0.61 (0.33) & 0.73 (0.46) & 0.32 (0.35) & 0.35 (0.36) & 0.67 (0.47) \\ 
        0.2 & 0.59 (0.34) & 0.63 (0.18) & 0.36 (0.14) & 0.19 (0.14) & 0.93 (0.11) \\ 
        0.4 & 0.57 (0.41) & 0.34 (0.25) & 0.34 (0.25) & 0.31 (0.11) & 0.86 (0.12) \\ 
        0.6 & 0.48 (0.29) & 0.42 (0.43) & 0.36 (0.43) & 0.34 (0.34) & 0.77 (0.18) \\ 
        0.8 & 0.52 (0.41) & 0.37 (0.35) & 0.30 (0.35) & 0.36 (0.13) & 0.75 (0.07) \\ 
        \bottomrule
    \end{tabular}
    \end{adjustbox}
\end{table*}

\begin{table*}[!htbp]
    \caption{\textbf{\textit{Effect of temperature:}} Analysis of temperature variation in various SELFIE based datasets with ICL (k=2) using CogVLM model.}
    \label{table:temperature_variations_cogvlm}
    \centering \small
    \begin{adjustbox}{max width=\textwidth}
    \begin{tabular}{lccccc}
        \toprule
        \textbf{Temp} & \textbf{BACE-V} & \textbf{BBBP-V} & \textbf{HIV-V} & \textbf{Clintox-V} & \textbf{Tox21-V} \\ 
        \midrule
        0.0 & 0.62 (0.44) & 0.51 (0.58) & 0.25 (0.29) & 0.32 (0.37) & 0.18 (0.14) \\ 
        0.2 & 0.58 (0.41) & 0.48 (0.54) & 0.28 (0.52) & 0.32 (0.31) & 0.14 (0.35) \\ 
        0.4 & 0.58 (0.41) & 0.42 (0.52) & 0.26 (0.48) & 0.33 (0.33) & 0.17 (0.34) \\ 
        0.6 & 0.53 (0.44) & 0.45 (0.48) & 0.25 (0.42) & 0.27 (0.31) & 0.14 (0.32) \\ 
        0.8 & 0.56 (0.39) & 0.37 (0.39) & 0.28 (0.49) & 0.25 (0.30) & 0.18 (0.32) \\ 
        \bottomrule
    \end{tabular}
    \end{adjustbox}
\end{table*}
\subsection{Impact of visual data}\label{ref:visual_data_classification}
Table \ref{table:llm_vlm} underscores the stark contrast in performance between Llama 2, a large language model, and its VLM counterpart, Llama Adapter v2 after ICL. Llama Adapter v2 also show substantial improvement post-finetuning.

\begin{table*}[!htbp]
    \caption{\textbf{\textit{Impact of visual data:}} First row shows Accuracy (F1-score) for ICL with language model Llama2, second row shows visual-language variant with improvement in performance, and third row demonstrates significant improvement in performance after finetuning.
    }
    \label{table:llm_vlm}
    \centering \small
    \begin{adjustbox}{}
    \begin{tabular}{lccccc}
        \toprule
        \textbf{Models} & \textbf{BACE-V} & \textbf{BBBP-V} & \textbf{HIV-V} & \textbf{ClinTox-V} & \textbf{Tox21-V} \\
        \midrule
          Llama 2 13B (ICL) & \textless
0.01(\textless
0.01) & 0.05(0.04) & 0.05(0.07) & 0.05(0.08) & \textless
0.01(\textless
0.01) \\
        \midrule
         Llama Adapter v2 7B (ICL) & 0.28(0.29) & 0.18(0.11) & 0.19(0.17) & 0.29(0.12) & 0.31(0.21) \\ 
         Llama Adapter v2 7B (LoRA) & 0.52(0.48) & 0.45(0.46) & 0.43(0.42) & 0.58(0.62) & 0.68(0.69) \\
        \bottomrule
    \end{tabular}
    \end{adjustbox}
\end{table*}

\begin{figure}[h]
\centerline{\includegraphics[width=0.6\columnwidth]{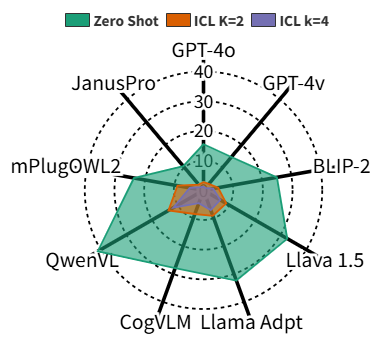}}
\caption{Radar plot comparing regression performance across various models (GPT-4v, GPT-4o, JansuPro, QwenVL, mPlugOWL2, BLIP-2, Llava 1.5 13B, CogVLM, and Llama Adapter v2 7B) averaged across ESOL, LD50, QM9, and PCQM4Mv2 datasets. The chart highlights Zero Shot (green) and Few Shot (k=2 in orange, k=4 in purple) capabilities.}
\label{fig:radar_plot_icl}
\end{figure}

\section{Regression: Further Analysis and Discussion}
\label{ref:Regression}

The main paper presents the results of regression tasks on ESOL, LD50, QM9, and PCQM4Mv2 for in-context learning (ICL) with \(k=2\) and finetuning. Here we provide additional results for \textbf{zero-shot} learning, detailed in Table \ref{table:performance_zero_shot_regression}, and \textbf{few-shot} learning with \(k=4\), shown in Table \ref{table:performance_few_shot_k4_regression}. Figure \ref{fig:radar_plot_icl} summarizes performance comparison across different models for regression tasks.

Furthermore, comprehensive evaluations on the QM9 dataset are included for 12 quantum mechanical targets under the \textbf{``all-together"} setting, where all targets are prompted simultaneously. These evaluations cover zero-shot learning (\(k=0\)), shown in Table \ref{table:performance_zero_shot_k0_QM9}; few-shot learning with \(k=2\), presented in Table \ref{table:performance_few_shot_k2_QM9}; and few-shot learning with \(k=4\), detailed in Table \ref{table:performance_few_shot_k4_QM9}.

\subsection{Effect of Finetuning}\label{ref:regression_finetuning}

The performance improvements achieved through LoRA-based finetuning are substantial across all regression tasks, as demonstrated in Table~\ref{table:performance_lora_regression}. BLIP-2 achieves the best overall performance with an average error of 1.925 across all datasets, excelling particularly on QM9-V (4.923 MAE) and PCQM4Mv2-V (0.235 MAE). CogVLM follows closely with an average error of 1.928, showing exceptional performance on ESOL-V with an RMSE of 1.102. Notably, Qwen VL demonstrates remarkable accuracy on LD50-V with a minimal MAE of 0.022. These finetuning results significantly outperform the in-context learning (ICL k=2) approaches presented in Table~3 of the main paper, where even the best-performing ICL model, Janus-Pro 7B, achieved only 2.52 average error. For example, BLIP-2's finetuned performance (1.925 average error) represents a 61.5\% improvement over its ICL performance (5.01 average error). Similarly, CogVLM improved from 8.50 to 1.928, and Qwen VL from 13.64 to 2.340. These dramatic improvements underscore the limitations of few-shot learning for molecular property prediction tasks and highlight the critical importance of task-specific parameter adaptation through finetuning.
\begin{table*}[!htbp]
    \caption{\textbf{\textit{Performance comparison after finetuning:}} Regression tasks. Error comparison of models finetuned using LoRA across different datasets. The best performing models are highlighted with bold text. Second best model performance are underlined.}
    \label{table:performance_lora_regression}
    \centering \footnotesize
    \begin{adjustbox}{max width=\textwidth,center}
    \begin{tabular}{lcccc|c}
    \toprule
       \textbf{Model} & \textbf{ESOL-V (RMSE)} & \textbf{LD50-V (MAE)} & \textbf{QM9-V (MAE)} & \textbf{PCQM4Mv2-V (MAE)} & \textbf{Average} \\
    \midrule
        BLIP-2                & 1.764  & 0.779  & \textbf{4.923}  & \textbf{0.235}   & \textbf{1.925} \\
        Llava 1.5 13B         & 2.229  & 0.193  & 5.193  & \underline{0.602}   & 2.554 \\
        Llama Adapter v2 7B   & 4.032  & 0.624  & 7.921  & 3.002   & 3.895 \\
        CogVLM                & \textbf{1.102}  & 0.592  & 5.221  & 0.795   & \underline{1.928} \\
        Qwen VL               & 2.192  & \textbf{0.022}  & \underline{5.021}  & 2.125   & 2.340 \\
        mPlugOWL2             & \underline{1.291}  & \underline{0.082}  & 8.029  & 1.621   & 2.756 \\
    \bottomrule
    \end{tabular}
    \end{adjustbox}
\end{table*}

\subsection{Effect of ICL Examples}\label{ref:regression_fewshot}
Tables \ref{table:performance_zero_shot_regression} and \ref{table:performance_few_shot_k4_regression} shows significant impact of in-context examples (k=4) versus zero-shot learning (k=0) across all models. GPT-4v maintains superior performance in both scenarios, ranking first with average metrics of 14.6485 and 1.04825 respectively, as shown in Table \ref{table:performance_zero_shot_regression} and Table \ref{table:performance_few_shot_k4_regression}. The introduction of examples leads to substantial error reduction, exemplified by GPT-4v's ESOL RMSE improving from 1.489 to 0.812, and its QM9 MAE decreasing from 45.296 to 2.503. Notably, the performance gap between models narrows with in-context examples, as evidenced by the reduction in average performance difference between best and worst models from 26.4815 (Table \ref{table:performance_zero_shot_regression}) to 10.69325 (Table \ref{table:performance_few_shot_k4_regression}), particularly in complex tasks like QM9 and PCQM4Mv2.

\begin{table*}[!htbp]
    \caption{\textbf{\textit{Performance comparison for zero-shot learning (k=0):}} MAE, RMSE of Multimodal LLMs for molecular property prediction based regression tasks.}
    \label{table:performance_zero_shot_regression}
    \centering \small
    \begin{adjustbox}{}
   \begin{tabular}{lccccc}
   \toprule
        \textbf{Model} & \textbf{ESOL (RMSE)} & \textbf{LD50 (MAE)} & \textbf{QM9 (MAE)} & \textbf{PCQM4Mv2 (MAE)} & \textbf{Average} \\
        \midrule
        GPT-4o    & \textbf{1.232}   & 12.31   & 47.126   & \textbf{1.41}   & 15.5195 \\ 
        GPT-4v    & 1.489   & 9.01   & \textbf{45.296}   & 2.799   & \textbf{14.6485} \\ 
        JanusPro 7B     & 1.997   & \textbf{8.124}   & 48.301   & 1.893   & 15.07875 \\ 
        BLIP-2     & 2.011   & 15.631   & 78.731   & 3.973   & 25.0865 \\ 
        Llava 1.5     & 18.198   & 14.952   & 93.182   & 3.967   & 32.57475 \\ 
        Llama Adapter    & 3.314   & 5.741   & 106.382   & 14.005   & 32.3605 \\ 
        CogVLM     & 2.093   & 23.769   & 80.766   & 4.939   & 27.89175 \\ 
        Qwen    & 4.119   & 24.388   & 116.353   & 19.66   & 41.13 \\ 
        mPlugOWL2   & 2.12   & 10.888   & 77.297   & 4.899   & 23.801 \\ 
        \bottomrule
    \end{tabular}
    \end{adjustbox}
\end{table*}
\begin{table*}[!htbp]
    \caption{\textbf{\textit{Performance comparison for few-shot learning (k=4):}} MAE, RMSE of Multimodal LLMs for molecular property prediction based regression tasks.}
    \label{table:performance_few_shot_k4_regression}
    \centering \small
    \begin{adjustbox}{}
    \begin{tabular}{lccccc}
    \toprule
        \textbf{Model} & \textbf{ESOL (RMSE)} & \textbf{LD50 (MAE)} & \textbf{QM9 (MAE)} & \textbf{PCQM4Mv2 (MAE)} & \textbf{Average} \\
        \midrule
        GPT-4o         & 0.867  & \textbf{0.614}  & 3.147   & 0.222   & 1.2125 \\ 
        GPT-4v        & 0.812 & 0.686 & \textbf{2.503} & \textbf{0.192}   & \textbf{1.048} \\ 
        JanusPro 7B      & \textbf{0.562} & 0.632  & 3.071  & 0.348   & 1.421 \\ 
        BLIP-2    & 1.289  & 0.696  & 10.339  & 0.882   & 3.3015 \\ 
        Llava 1.5   & 4.361  & 0.709  & 20.157  & 0.883   & 6.5275 \\ 
        Llama Adapter  & 2.309  & 2.431  & 19.840  & 3.809   & 7.09725 \\ 
        CogVLM    & 1.225  & 0.815  & 15.662  & 0.805   & 4.62675 \\ 
        Qwen     & 3.332  & 0.839  & 33.534  & 9.261   & 11.7415 \\ 
        mPlugOWL2  & 1.416  & 0.741  & 14.998  & 1.692   & 4.71175 \\ 
        \bottomrule
    \end{tabular}
    \end{adjustbox}
\end{table*}
\subsection{Analysis of QM9 Multi-Target Prediction}\label{ref:qm9-multi-target}
Tables \ref{table:performance_zero_shot_k0_QM9}, \ref{table:performance_few_shot_k2_QM9}, and \ref{table:performance_few_shot_k4_QM9} present the performance comparison for simultaneous prediction of all 12 QM9 molecular properties. In the zero-shot setting (Table \ref{table:performance_zero_shot_k0_QM9}), GPT-4o and GPT-4v demonstrate superior performance with average MAEs of 47.1264 and 45.2964 respectively. The addition of in-context examples (k=2, k=4) significantly improves prediction accuracy across all models, with GPT-4v achieving the best average MAE of 2.5028 at k=4. Notably, both GPT-4o and GPT-4v show consistent performance across different molecular properties, maintaining their superiority even in this challenging multi-target prediction scenario.
\begin{table*}[!htbp]
    \caption{\textbf{\textit{Performance comparison for zero-shot learning (k=0):}} MAE of Multimodal LLMs for molecular property prediction based regression tasks on QM9 Dataset.}
    \label{table:performance_zero_shot_k0_QM9}
    \centering \small
    \begin{adjustbox}{width=\textwidth}
    \begin{tabular}{lccccccccccccc}
    \toprule
        \textbf{Model} & \rotatebox{90}{\textbf{QM9(Alpha)}} & \rotatebox{90}{\textbf{QM9(Gap)}} & \rotatebox{90}{\textbf{QM9(Homo)}} & \rotatebox{90}{\textbf{QM9(Lumo)}} & \rotatebox{90}{\textbf{QM9(Mu)}} & \rotatebox{90}{\textbf{QM9(CV)}} & \rotatebox{90}{\textbf{QM9(G298)}} & \rotatebox{90}{\textbf{QM9(H298)}} & \rotatebox{90}{\textbf{QM9(r2)}} & \rotatebox{90}{\textbf{QM9(u298)}} & \rotatebox{90}{\textbf{QM9(u0)}} & \rotatebox{90}{\textbf{QM9(zpve)}} & \rotatebox{90}{\textbf{QM9(Avg)}} \\
        \midrule
        GPT-4o          & \textbf{11.0855} & \textbf{1.8364} & \textbf{1.0287} & 3.0245 & \textbf{2.6813} & 6.7424 & \textbf{85.0988} & \textbf{85.0906} & \textbf{195.5142} & \textbf{85.0902} & \textbf{85.0929} & 3.2317 & \textbf{47.1264} \\ 
        GPT-4v          & 14.925  & 2.4481  & 1.9924  & \textbf{2.129} & 3.129  & 6.1294 & 88.924  & 120.449 & 145.9812 & 66.2498 & 88.9192 & \textbf{2.2804} & 45.2964 \\ 
         JanusPro 7B & 12.921 & 3.8238 & 1.8299 & 2.921 & 3.109 & 4.291 & 97.921 & 91.842 & 170.829 & 73.289 & 111.924 & 4.921 & 48.301 \\ 
        BLIP-2          & 47.902  & 3.920   & 2.220   & 2.844  & 11.294 & 59.201 & 122.842 & 129.912 & 255.901  & 102.120  & 194.201 & 12.422  & 78.7316 \\ 
                Llava  1.5         & 108.449 & 32.923  & 21.382  & 17.994 & 17.544 & 36.686 & 173.561 & 126.244 & 209.361  & 186.335  & 157.421 & 30.285  & 93.1821 \\ 
        Llama-Adapter   & 175.989 & 13.822  & 11.939  & 15.679 & 12.525 & 79.738 & 207.088 & 130.005 & 232.172  & 165.611  & 201.434 & 30.584  & 106.3822 \\ 
        CogVLM          & 93.191  & 2.119   & 1.4757  & 2.997  & 9.784  & 45.534 & 137.238 & 184.106 & 232.069  & 132.701  & 120.716 & 7.2616  & 80.7660 \\ 
        Qwen            & 184.191 & 8.201   & 7.772   & 9.009  & 46.111 & 137.302 & 240.019 & 129.090 & 220.322  & 172.828  & 197.466 & 43.923  & 116.3528 \\ 
        mPlugOWL2       & 72.706  & 3.676   & 4.280   & 2.457  & 5.673  & 31.363 & 157.058 & 157.078 & 214.541  & 100.495  & 121.920 & 56.315  & 77.2968 \\ 
        \bottomrule
    \end{tabular}
    \end{adjustbox}
\end{table*}

\begin{table*}[!htbp]
    \caption{\textbf{\textit{Performance comparison for few-shot learning (k=2):}} MAE of Multimodal LLMs for molecular property prediction based regression tasks on QM9 Dataset.}
    \label{table:performance_few_shot_k2_QM9}
    \centering \small
    \begin{adjustbox}{width=\textwidth}
    \begin{tabular}{lccccccccccccc}
    \toprule
        \textbf{Model} & \rotatebox{90}{\textbf{QM9(Alpha)}} & \rotatebox{90}{\textbf{QM9(Gap)}} & \rotatebox{90}{\textbf{QM9(Homo)}} & \rotatebox{90}{\textbf{QM9(Lumo)}} & \rotatebox{90}{\textbf{QM9(Mu)}} & \rotatebox{90}{\textbf{QM9(CV)}} & \rotatebox{90}{\textbf{QM9(G298)}} & \rotatebox{90}{\textbf{QM9(H298)}} & \rotatebox{90}{\textbf{QM9(r2)}} & \rotatebox{90}{\textbf{QM9(u298)}} & \rotatebox{90}{\textbf{QM9(u0)}} & \rotatebox{90}{\textbf{QM9(zpve)}} & \rotatebox{90}{\textbf{QM9(Avg)}} \\
        \midrule
        GPT-4o          & \textbf{2.32} & \textbf{0.2126} & \textbf{0.3113} & \textbf{0.2212} & 0.9787 & \textbf{0.857} & \textbf{17.9428} & \textbf{17.9434} & \textbf{22.948} & \textbf{17.9435} & \textbf{17.9432} & 0.9126 & \textbf{8.3779} \\ 
        GPT-4v          & 3.1 & 0.484 & 0.396 & 0.293 & \textbf{0.968} & 0.998 & 18.021 & 18.022 & 24.405 & 18.022 & 18.021 & \textbf{0.725} & 8.6217 \\ 
        JanusPro 7B & 2.892 & 0.429 & 0.3329 & 0.3392 & 0.792 & 0.729 & 18.291 & 18.728 & 21.882 & 19.821 & 17.211 & 0.884 & 8.5276\\ 
        BLIP-2          & 3.401 & 0.713 & 0.891 & 0.629 & 2.14 & 2.129 & 22.092 & 21.921 & 85.239 & 29.912 & 22.12 & 0.982 & 16.014 \\ 
         Llava 1.5          & 17.472 & 2.191 & 1.991 & 1.629 & 7.737 & 3.516 & 26.653 & 68.502 & 100.307 & 56.087 & 36.873 & 1.024 & 26.9985 \\ 
        Llama-adapter   & 17.316 & 1.621 & 2.091 & 1.032 & 4.554 & 5.011 & 39.585 & 45.626 & 130.37 & 58.654 & 29.255 & 2.02 & 28.0946 \\ 
        CogVLM          & 7.6465 & 0.91 & 1.012 & 0.876 & 3.613 & 2.799 & 21.292 & 27.702 & 174.554 & 45.178 & 23.74 & 0.822 & 25.8454 \\ 
                QwenVLM            & 13.563 & 2.03 & 1.293 & 1.075 & 10.587 & 3.563 & 130.992 & 75.385 & 134.581 & 55.007 & 36.941 & 2.036 & 38.9211 \\ 
        mPlugOWL2       & 10.336 & 1.832 & 1.432 & 1.71 & 2.685 & 3.233 & 23.091 & 37.263 & 189.806 & 52.405 & 27.379 & 0.829 & 29.333
 \\

        \bottomrule
    \end{tabular}
    \end{adjustbox}
\end{table*}

\begin{table*}[!htbp]
    \caption{\textbf{\textit{Performance comparison for few-shot learning (k=4):}} MAE of Multimodal LLMs for molecular property prediction based regression tasks on QM9 Dataset.}
    \label{table:performance_few_shot_k4_QM9}
    \centering \small
    \begin{adjustbox}{width=\textwidth}
    \begin{tabular}{lccccccccccccc}
    \toprule
        \textbf{Model} & \rotatebox{90}{\textbf{QM9(Alpha)}} & \rotatebox{90}{\textbf{QM9(Gap)}} & \rotatebox{90}{\textbf{QM9(Homo)}} & \rotatebox{90}{\textbf{QM9(Lumo)}} & \rotatebox{90}{\textbf{QM9(Mu)}} & \rotatebox{90}{\textbf{QM9(CV)}} & \rotatebox{90}{\textbf{QM9(G298)}} & \rotatebox{90}{\textbf{QM9(H298)}} & \rotatebox{90}{\textbf{QM9(r2)}} & \rotatebox{90}{\textbf{QM9(u298)}} & \rotatebox{90}{\textbf{QM9(u0)}} & \rotatebox{90}{\textbf{QM9(zpve)}} & \rotatebox{90}{\textbf{QM9(Avg)}} \\
        \midrule
        GPT-4o          & \textbf{1.28} & \textbf{0.0184} & \textbf{0.0137} & \textbf{0.0053} & \textbf{0.2787} & 0.257 & 3.9572 & 3.9561 & 20.052 & 3.956 & 3.9563 & 0.0334 & 3.1470 \\ 
        GPT-4v          & 1.56 & 0.0244 & 0.0383 & 0.0069 & 0.9664 & \textbf{0.207} & \textbf{3.0218} & \textbf{3.0214} & \textbf{15.1194} & \textbf{3.0215} & 3.0215 & \textbf{0.0251} & \textbf{2.5028} \\ 
       JanusPro 7B & 1.992 & 0.0892 & 0.0782 & 0.0098 & 0.389 & 0.108 & 5.208 & 4.592 & 17.229 & 4.092 & \textbf{2.981} & 0.0782 & 3.0705 \\
        BLIP-2          & 2.992 & 0.064 & 0.192 & 0.023 & 1.102 & 0.942 & 9.28 & 19.029 & 73.923 & 9.284 & 7.128 & 0.103 & 10.3385 \\ 
        CogVLM          & 6.9543 & 0.166 & 0.112 & 0.196 & 1.644 & 2.912 & 16.482 & 19.597 & 98.439 & 18.502 & 22.844 & 0.0919 & 15.6617 \\ 
        mPlugOWL2       & 8.823 & 1.016 & 1.011 & 1.489 & 1.556 & 2.746 & 14.191 & 22.711 & 88.797 & 15.687 & 21.826 & 0.126 & 14.9983 \\ 
        llava           & 10.064 & 0.362 & 0.221 & 0.092 & 1.76 & 3.014 & 21.824 & 65.123 & 80.81 & 28.139 & 30.044 & 0.426 & 20.1566 \\ 
        llama-adapter   & 15.961 & 1.129 & 1.071 & 0.902 & 2.65 & 4.741 & 25.978 & 27.412 & 93.616 & 37.095 & 26.511 & 1.019 & 19.8404 \\ 
        qwen            & 12.988 & 1.203 & 1.102 & 1.091 & 2.813 & 2.988 & 97.619 & 69.377 & 127.947 & 51.125 & 33.136 & 1.018 & 33.5339 \\ 
        \bottomrule
    \end{tabular}
    \end{adjustbox}
\end{table*}
\section{Molecular Description: Further Analysis}\label{ref:molecular_description}
We conducted a comprehensive evaluation of molecular description capabilities across multiple models and settings. This analysis examines zero-shot performance, the effects of in-context learning (ICL) with varying numbers of examples, and the impact of Chain of Thought (CoT) prompting on description quality. All experiments use the same evaluation metrics: BLEU-2, BLEU-4, ROUGE-1, ROUGE-2, ROUGE-L, and METEOR, with average scores reported for concise comparison.
\subsection{Zero-Shot Evaluation}\label{ref:description_zero_shot}
We conducted experiments evaluating molecular description capabilities in a zero-shot setting. GPT-4o demonstrates superior performance across all metrics, achieving the highest average score of 27.541 (Table \ref{table:performance_molecular_description_zero_shot}). GPT-4v follows with a notable performance gap but consistent profile (24.556 average). Among the remaining models, JanusPro and Llava 1.5 13B form the second tier (19.642 and 18.140 average, respectively), followed by CogVLM and mPlugOWL2 showing comparable capabilities (16.158 and 16.004). BLIP-2 and Qwen VL deliver similar mid-to-low range performance, while Llama Adapter v2 7B struggles significantly with this task (10.048 average). These results suggest that general-purpose models with extensive pretraining currently maintain substantial advantages in zero-shot molecular understanding and description tasks.
\begin{table*}[!htbp]
    \caption{\textbf{\textit{Molecular description performance in zero-shot setting:}} Comparison of models evaluated on molecular description task without finetuning. The best performing models are highlighted with bold text.}
    \label{table:performance_molecular_description_zero_shot}
    \centering \footnotesize
    \begin{adjustbox}{max width=\textwidth,center}
    \begin{tabular}{lccccccc}
        \toprule
        \textbf{Models} & \textbf{BLEU-2 $\uparrow$} & \textbf{BLEU-4 $\uparrow$} & \textbf{ROUGE-1 $\uparrow$} & \textbf{ROUGE-2 $\uparrow$} & \textbf{ROUGE-L $\uparrow$} & \textbf{METEOR $\uparrow$} & \textbf{Average $\uparrow$} \\
        \midrule
        GPT-4o & \textbf{27.853} & \textbf{25.491} & \textbf{29.376} & \textbf{26.184} & \textbf{28.729} & \textbf{27.615} & \textbf{27.541} \\
        GPT-4v & 24.837 & 23.492 & 25.876 & 24.184 & 25.329 & 23.615 & 24.556 \\
        BLIP-2 & 12.610 & 11.780 & 12.480 & 12.150 & 12.390 & 11.940 & 12.225 \\
        CogVLM & 16.753 & 15.292 & 16.876 & 15.684 & 16.529 & 15.815 & 16.158 \\
        mPlugOWL2 & 16.621 & 15.220 & 16.432 & 15.837 & 16.286 & 15.629 & 16.004 \\
        Llava 1.5 13B & 18.662 & 17.544 & 18.470 & 18.047 & 18.340 & 17.774 & 18.140 \\
        Llama Adapter v2 7B & 10.332 & 9.723 & 10.235 & 9.985 & 10.173 & 9.841 & 10.048 \\
        Qwen VL & 14.497 & 13.634 & 14.360 & 14.008 & 14.267 & 13.802 & 14.095 \\
        JanusPro & 19.753 & 18.492 & 20.876 & 19.284 & 20.529 & 18.915 & 19.642 \\
        \bottomrule
    \end{tabular}
    \end{adjustbox}
    \vspace{-10pt}
\end{table*}

\subsection{Effect of ICL Examples}\label{ref:description_icl}
Tables \ref{table:performance_molecular_description_few_shot_k2} and \ref{table:performance_molecular_description_few_shot_k4} demonstrate significant performance improvements when increasing from 2-shot to 4-shot learning. GPT-4v leads in the 2-shot setting with the highest average score of 43.400, while GPT-4o achieves superior results in the 4-shot setting with an average of 59.727, showing a remarkable improvement of 17.6 percentage points over its 2-shot performance.
All models exhibit consistent gains when provided with additional examples, with larger models generally demonstrating better utilization of in-context examples. JanusPro maintains strong performance in both settings, while smaller models like Llama Adapter v2 7B show more modest improvements.
\begin{table*}[!htbp]
    \caption{\textbf{\textit{Molecular description performance in few-shot setting (k=2):}} Comparison of models evaluated on molecular description task with 2-shot learning. The best performing models are highlighted with bold text.}
    \label{table:performance_molecular_description_few_shot_k2}
    \centering \footnotesize
    \begin{adjustbox}{max width=\textwidth,center}
    \begin{tabular}{lccccccc}
        \toprule
        \textbf{Models} & \textbf{BLEU-2 $\uparrow$} & \textbf{BLEU-4 $\uparrow$} & \textbf{ROUGE-1 $\uparrow$} & \textbf{ROUGE-2 $\uparrow$} & \textbf{ROUGE-L $\uparrow$} & \textbf{METEOR $\uparrow$} & \textbf{Average $\uparrow$} \\
        \midrule
        GPT-4o & 43.330 & 40.180 & 42.870 & 42.070 & 42.670 & 41.740 & 42.143 \\
        GPT-4v & \textbf{43.720} & \textbf{43.020} & \textbf{43.620} & \textbf{43.310} & \textbf{43.580} & \textbf{43.150} & \textbf{43.400} \\
        BLIP-2 & 30.265 & 29.273 & 30.136 & 29.719 & 30.032 & 29.546 & 29.829 \\
        CogVLM & 31.951 & 30.552 & 31.762 & 31.167 & 31.616 & 30.959 & 31.335 \\
        mPlugOWL2 & 26.664 & 24.795 & 26.390 & 25.619 & 26.184 & 25.359 & 25.835 \\
        Llava 1.5 13B & 29.818 & 28.617 & 29.653 & 29.137 & 29.526 & 28.951 & 29.284 \\
        Llama Adapter v2 7B & 21.619 & 20.465 & 21.469 & 21.142 & 21.377 & 20.982 & 21.176 \\
        Qwen VL & 31.062 & 29.051 & 30.817 & 29.923 & 30.600 & 29.583 & 30.173 \\
        JanusPro & 38.286 & 37.061 & 38.153 & 37.618 & 38.021 & 37.262 & 37.734 \\
        \bottomrule
    \end{tabular}
    \end{adjustbox}
    \vspace{-10pt}
\end{table*}

\begin{table*}[!htbp]
    \caption{\textbf{\textit{Molecular description performance in few-shot setting (k=4):}} Comparison of models evaluated on molecular description task with 4-shot learning. The best performing models are highlighted with bold text.}
    \label{table:performance_molecular_description_few_shot_k4}
    \centering \footnotesize
    \begin{adjustbox}{max width=\textwidth,center}
    \begin{tabular}{lccccccc}
        \toprule
        \textbf{Models} & \textbf{BLEU-2 $\uparrow$} & \textbf{BLEU-4 $\uparrow$} & \textbf{ROUGE-1 $\uparrow$} & \textbf{ROUGE-2 $\uparrow$} & \textbf{ROUGE-L $\uparrow$} & \textbf{METEOR $\uparrow$} & \textbf{Average $\uparrow$} \\
        \midrule
        GPT-4o & \textbf{61.310} & \textbf{58.690} & \textbf{60.490} & \textbf{59.220} & \textbf{60.060} & \textbf{58.590} & \textbf{59.727} \\
        GPT-4v & 55.780 & 54.430 & 55.580 & 54.930 & 55.460 & 54.650 & 55.138 \\
        BLIP-2 & 36.064 & 34.530 & 35.896 & 35.227 & 35.728 & 35.049 & 35.416 \\
        CogVLM & 40.285 & 39.062 & 40.152 & 39.618 & 40.020 & 39.262 & 39.733 \\
        mPlugOWL2 & 33.774 & 32.299 & 33.614 & 32.969 & 33.454 & 32.792 & 33.150 \\
        Llava 1.5 13B & 40.775 & 39.507 & 40.638 & 40.083 & 40.502 & 39.715 & 40.203 \\
        Llama Adapter v2 7B & 32.063 & 31.075 & 31.947 & 31.456 & 31.857 & 31.298 & 31.616 \\
        Qwen VL & 40.285 & 39.062 & 40.152 & 39.618 & 40.020 & 39.262 & 39.733 \\
        JanusPro & 50.574 & 48.696 & 50.373 & 49.599 & 50.104 & 49.098 & 49.741 \\
        \bottomrule
    \end{tabular}
    \end{adjustbox}
    \vspace{-10pt}
\end{table*}
\subsection{Chain of Thought Prompting}\label{ref:description_cot}
Table \ref{table:performance_molecular_description_cot} illustrates the impact of Chain of Thought (CoT) prompting on molecular description tasks. GPT-4o achieves superior performance across all metrics with an average score of 61.494, slightly outperforming GPT-4v (59.549). Both models demonstrate strong capabilities when encouraged to reason step-by-step. JanusPro maintains its position as the third-best performer with an average score of 53.703, showing considerable potential among non-commercial models. The remaining models show varying degrees of effectiveness with CoT prompting, with Llava and CogVLM achieving similar performance (43.393 and 42.878, respectively). Notably, when compared to few-shot learning results, CoT prompting appears to further enhance model performance, particularly for larger models, suggesting that structured reasoning approaches are beneficial for molecular description tasks.
\begin{table*}[!htbp]
   \caption{\textbf{\textit{Molecular description performance in Chain-of-thought setting:}} Comparison of models evaluated on molecular description task with Chain of Thought (CoT). The best performing models are highlighted with bold text.}
    \label{table:performance_molecular_description_cot}
    \centering \footnotesize
    \begin{adjustbox}{max width=\textwidth,center}
    \begin{tabular}{lccccccc}
        \toprule
        \textbf{Models} & \textbf{BLEU-2 $\uparrow$} & \textbf{BLEU-4 $\uparrow$} & \textbf{ROUGE-1 $\uparrow$} & \textbf{ROUGE-2 $\uparrow$} & \textbf{ROUGE-L $\uparrow$} & \textbf{METEOR $\uparrow$} & \textbf{Average $\uparrow$} \\
        \midrule
        GPT-4o & \textbf{62.162} & \textbf{60.743} & \textbf{62.406} & \textbf{61.151} & \textbf{62.006} & \textbf{60.495} & \textbf{61.494} \\
        GPT-4v & 60.242 & 58.784 & 60.026 & 59.324 & 59.897 & 59.022 & 59.549 \\
        BLIP2 & 38.708 & 37.092 & 38.530 & 37.825 & 38.344 & 37.638 & 38.023 \\
        CogVLM & 43.467 & 42.187 & 43.324 & 42.745 & 43.182 & 42.361 & 42.878 \\
        mPlugOWL & 36.475 & 34.883 & 36.303 & 35.606 & 36.130 & 35.416 & 35.802 \\
        Llava & 43.996 & 42.668 & 43.849 & 43.249 & 43.702 & 42.892 & 43.393 \\
        Llama Adapter & 34.628 & 33.561 & 34.503 & 33.972 & 34.405 & 33.802 & 34.145 \\
        Qwen & 43.387 & 42.187 & 43.324 & 42.745 & 43.182 & 42.403 & 42.871 \\
        Janus & 54.571 & 52.591 & 54.352 & 53.567 & 54.112 & 53.026 & 53.703 \\
        \bottomrule
    \end{tabular}
    \end{adjustbox}
    \vspace{-10pt}
\end{table*}
\section{Contrastive Learning for Vision Encoders}\label{ref:contrastive_learning}
We explore two contrastive learning strategies for enhancing the vision encoder's ability to capture molecular structural information: augmentation-based and Tanimoto similarity-based approaches. This additional loss is used with LoRA finetuning. We experimente with BLIP-2 considering its better performance across all tasks. The motivation is to enable the vision encoder to learn more discriminative representations of molecular structures by leveraging either image transformations or chemical similarity relationships. The performance of these contrastive learning approaches across multiple molecular datasets is summarized in the main paper, while the detailed metrics on the molecular description task are presented in Table~\ref{table:performance_molecular_description_contrastive}.
\begin{table*}[!htbp]
    \caption{\textbf{\textit{Molecular description performance using Contrastive Learning}} Comparison of BLIP2 evaluated on molecular description task with Augmentation (Aug) and Tanimoto Augmentation (T-Aug). }
    \label{table:performance_molecular_description_contrastive}
    \centering \footnotesize
    \begin{adjustbox}{max width=\textwidth,center}
    \begin{tabular}{lccccccc}
        \toprule
        \textbf{Models} & \textbf{BLEU-2 $\uparrow$} & \textbf{BLEU-4 $\uparrow$} & \textbf{ROUGE-1 $\uparrow$} & \textbf{ROUGE-2 $\uparrow$} & \textbf{ROUGE-L $\uparrow$} & \textbf{METEOR $\uparrow$} & \textbf{Average $\uparrow$} \\
        \midrule
        Lora & 59.062 & 58.034 & 58.933 & 58.466 & 58.889 & 58.185 & 58.595 \\
        Aug & 61.530 & 60.307 & 61.398 & 60.863 & 61.266 & 60.507 & 60.979 \\
        T-Aug & \textbf{64.176} & \textbf{63.192} & \textbf{64.072} & \textbf{63.641} & \textbf{63.966} & \textbf{63.352} & \textbf{63.733} \\
        \bottomrule
    \end{tabular}
    \end{adjustbox}
    \vspace{-10pt}
\end{table*}

\subsection{Augmentation-based contrastive learning}\label{ref:contrastive_augmentation}
We generate multiple views of the same molecule using a set of image transformations including rotations (at angles 45°, 90°, 135°, 180°, 225°, 270°, and 315°), vertical and horizontal flips, solarization, posterization, and auto-contrast adjustments. For each molecule image, we randomly apply two transformations to create positive pairs for contrastive learning.

\begin{figure}[htbp]
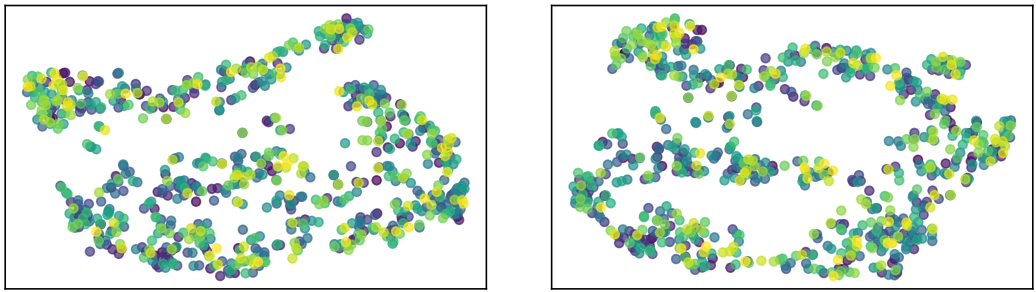

    \centering
    \begin{minipage}{0.48\textwidth}
        \centering
        \includesvg[width=\textwidth]{test_vision_encoder_tsne_augmented.svg}
    \end{minipage}%
    \hfill
    \begin{minipage}{0.48\textwidth}
        \centering
        \includesvg[width=\textwidth]{test_multimodal_fusion_tsne_augmented.svg}
    \end{minipage}
\caption{\textbf{\textit{Analyzing visual features:}} The two plots show t-SNE visualizations of visual encodings of BLIP-2 before and after cross-modal fusion respectively using augmentative technique.}

    \label{fig:tsne-augmentation}
\end{figure}

Figure \ref{fig:tsne-augmentation} illustrates the analysis of our augmentation-based approach through t-SNE visualizations of the visual encodings. While our main paper demonstrates the superior performance of contrastive learning using Tanimoto similarity, here we present results from image augmentation techniques for comparison. The left plot shows the visual encodings from BLIP-2 before cross-modal fusion, with clusters exhibiting significant overlap. The right plot displays the representations after cross-modal fusion, where the clusters become more distinguishable but still less defined than those achieved with Tanimoto similarity methods discussed in the main paper. This comparative analysis confirms that Tanimoto similarity-based approaches provide better molecular structure differentiation than augmentation-based techniques alone, particularly for chemical structure representation tasks.

\subsection{Tanimoto similarity-based contrastive learning}\label{ref:contrastive_tanimoto}
Rather than using augmented views, we leverage chemical similarity to define positive pairs. For each molecule, we identify three structurally similar molecules with Tanimoto similarity scores $>$0.85 to serve as positive examples. This approach ensures that the model learns from meaningful chemical relationships rather than artificial transformations.

\subsection{Overall loss function}\label{ref:contrastive_loss_function}

For both approaches, we implement a contrastive loss based on NT-Xent (Normalized Temperature-scaled Cross Entropy) as used in SimCLR. The fundamental principle behind this loss function is to learn discriminative molecular representations by pulling together embeddings of similar molecules (positive pairs) while pushing apart embeddings of dissimilar molecules (negative pairs) in the representation space.

The contrastive loss is mathematically defined as:
\begin{equation}
\mathcal{L}_{\text{contrastive}} = -\frac{1}{2N}\sum_{i=1}^{N}\log\frac{\exp(\text{sim}(z_i, z_j)/\tau)}{\sum_{k=1}^{2N}\mathbf{1}_{[k\neq i]}\exp(\text{sim}(z_i, z_k)/\tau)}
\end{equation}

The key components of this formulation include $z_i$ and $z_j$, which represent the normalized embeddings of a positive pair obtained from the vision encoder after processing molecular images. The similarity function $\text{sim}(z_i, z_j) = \frac{z_i^T z_j}{\|z_i\|\|z_j\|}$ denotes the cosine similarity between two embeddings, while $\tau = 0.5$ is the temperature parameter that controls the concentration of the distribution around positive pairs. The batch size is represented by $N$, resulting in $2N$ total samples when considering both elements of each positive pair, and $\mathbf{1}_{[k\neq i]}$ is an indicator function that excludes the case where $k=i$ to prevent self-comparison.

The loss function operates by computing the probability that embedding $z_i$ is most similar to its positive counterpart $z_j$ compared to all other embeddings in the batch. The numerator $\exp(\text{sim}(z_i, z_j)/\tau)$ represents the similarity between the positive pair, while the denominator sums over all possible negative pairs within the batch, creating a softmax-like normalization that encourages the model to distinguish between related and unrelated molecular structures.

The temperature parameter $\tau$ plays a crucial role in controlling the learning dynamics. A lower temperature such as our chosen value of 0.5 creates sharper distributions, making the model more sensitive to small differences in similarity scores and encouraging tighter clustering of positive pairs. This helps the encoder learn more discriminative features that are essential for downstream molecular property prediction across classification tasks, regression tasks, and molecular description tasks.

During training, this contrastive loss is combined with the task-specific loss for the target dataset:
\begin{equation}
\mathcal{L}_{\text{total}} = \mathcal{L}_{\text{task}} + \lambda \mathcal{L}_{\text{contrastive}}
\end{equation}

Here, $\mathcal{L}_{\text{task}}$ represents the primary loss for the specific molecular property prediction task, which varies depending on whether we are performing classification (cross-entropy loss), regression (mean squared error or mean absolute error), or molecular description (sequence generation loss). The weighting parameter $\lambda$ balances the contribution of the contrastive learning objective with the task-specific objective. This combined loss function is optimized during LoRA finetuning, allowing the vision encoder to simultaneously learn task-specific features for classification, regression, and molecular description tasks while maintaining the ability to distinguish between different molecular structures through contrastive learning. The integration of contrastive learning with task-specific objectives enables the model to develop robust molecular representations that generalize well across diverse downstream applications in molecular property prediction.

\section{Prompt Examples}
\label{prompt_examples}

\begin{figure*}[h]
\centerline{\includegraphics[width=0.8\textwidth]{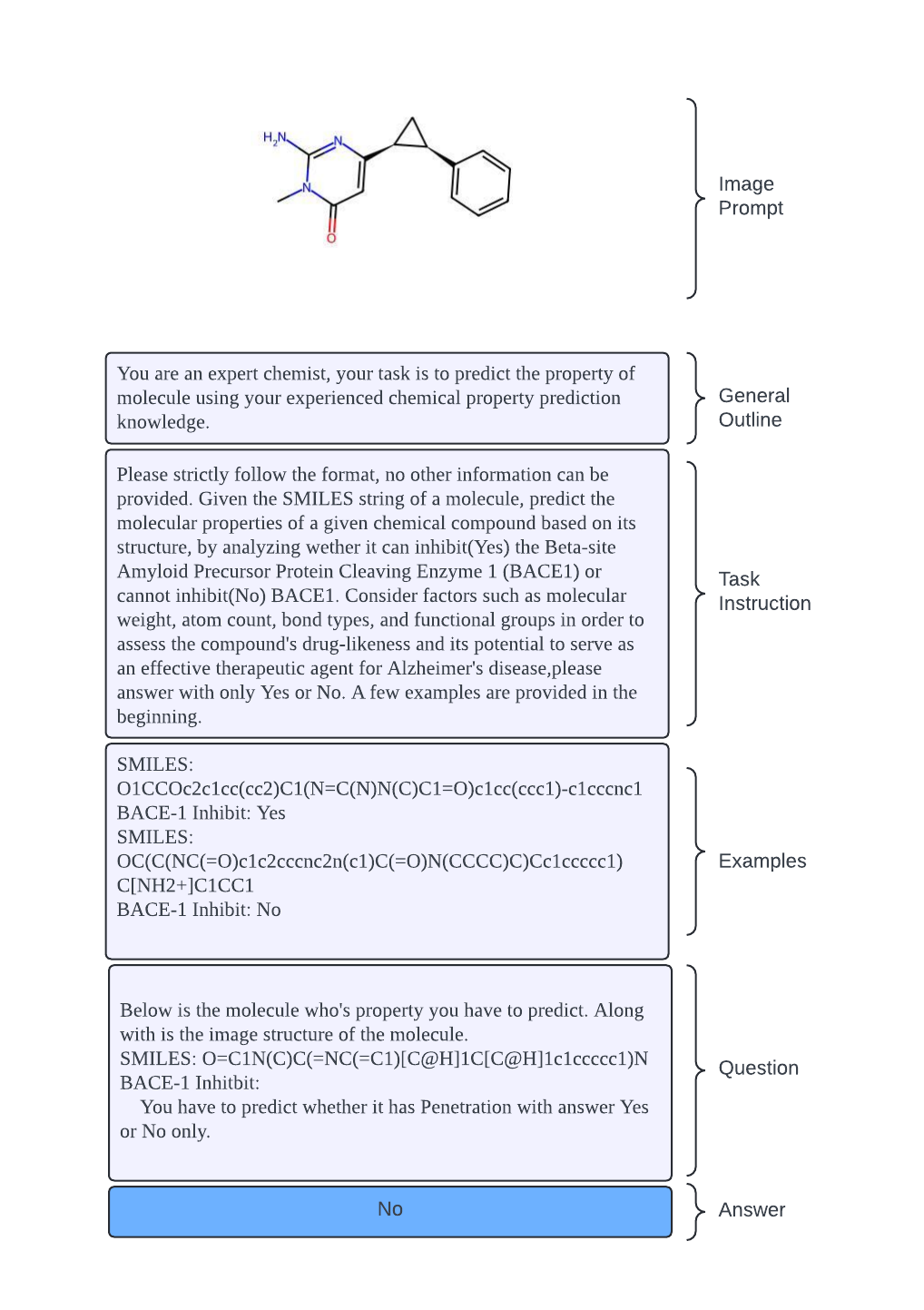}}
\caption{\textbf{\textit{Sample prompt for BACE-V:}} A general outline is provided at first followed by set of instructions to be more specific about the task. The task is explained briefly and expected output is stated. In our case it should be Yes/No. This includes ICL examples with k=2 (No of samples). With this the main question is asked. The chemical compound who's property is to be expected is represented in its molecular structure created using RDKIT which goes along with the text input.}
\label{fig:bacesmileprompt}
\end{figure*}

\begin{figure*}[h]
\centerline{\includegraphics[width=0.75\textwidth]{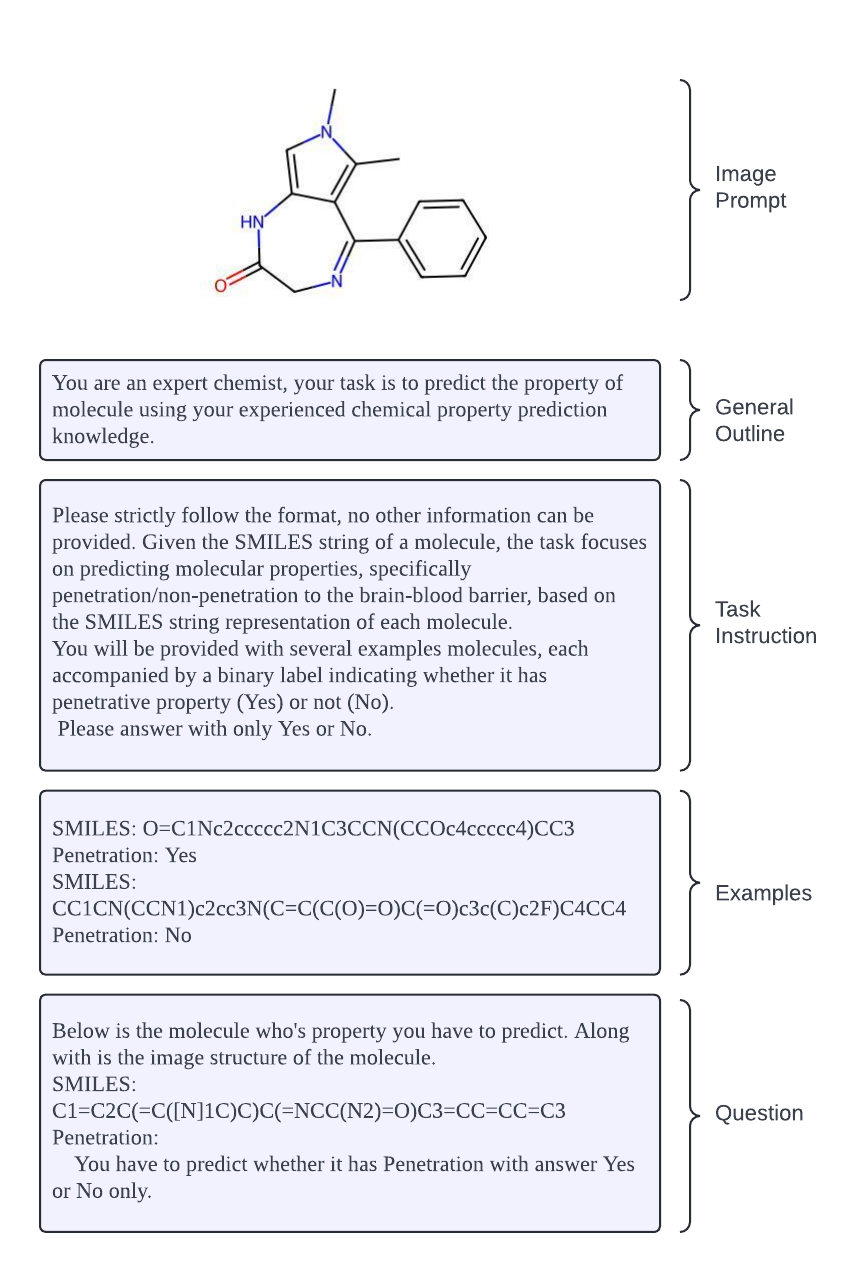}}
\caption{\textbf{\textit{Example prompt:}} The figure presents a task designed for predicting molecular properties, specifically penetration through the blood-brain barrier (BBBP-V dataset), using the SMILES string representation. The general outline and specific instructions detail the expected binary output (Yes/No). Two example molecules are provided to illustrate the task, followed by the main question, which includes the SMILES string and structure of the target molecule, generated with RDKit.}
\label{fig:bbbpsmileprompt}
\end{figure*}

\begin{figure*}[h]
\centerline{\includegraphics[width=0.8\textwidth]{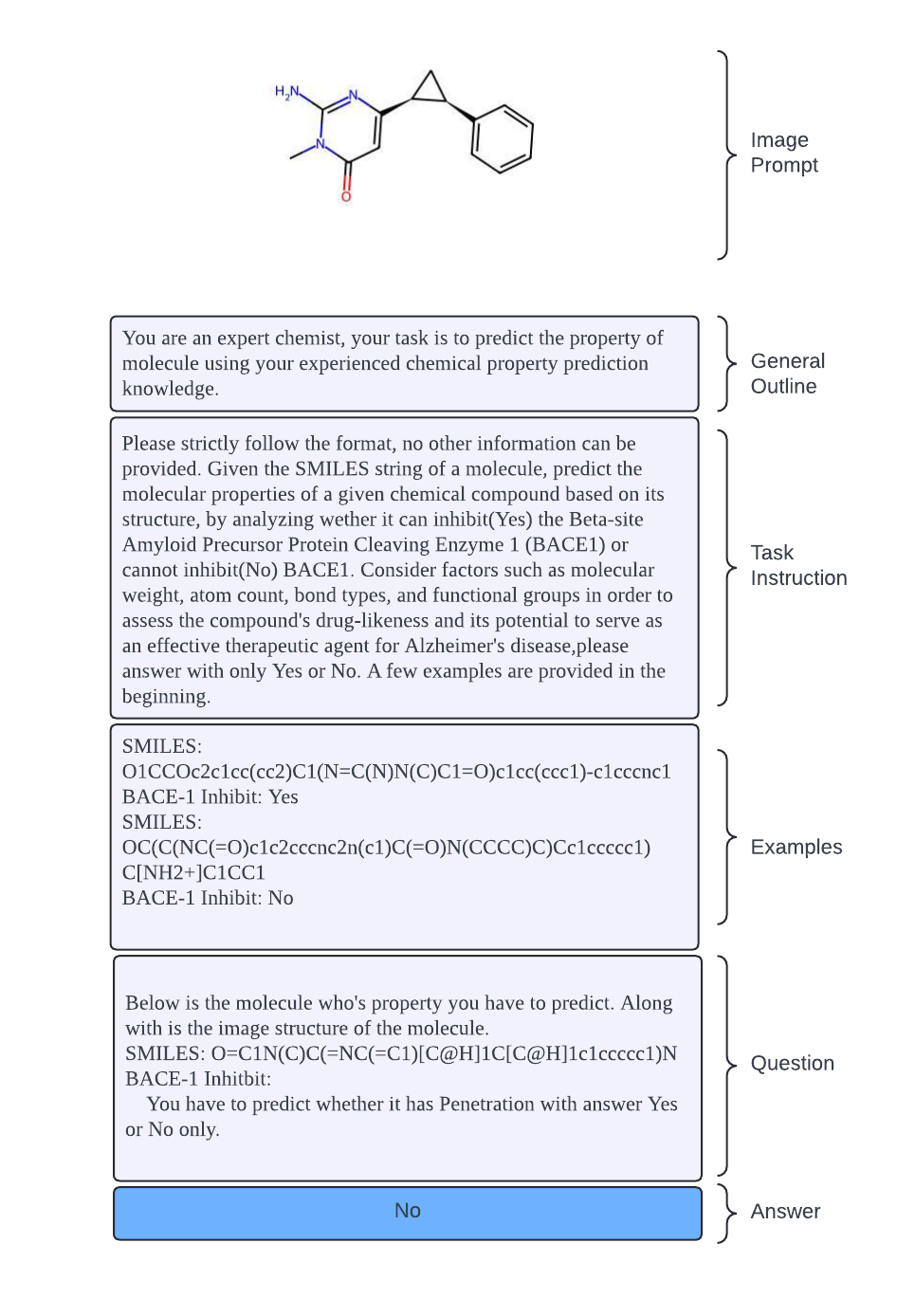}}
\caption{\textbf{\textit{Example prompt:}} The figure outlines a task for predicting the ability of molecules to inhibit HIV replication (HIV-V dataset), based on their SMILES string representation. The general outline and specific instructions require a binary output (Yes/No). Example molecules are provided to illustrate the task, followed by the main question, which includes the SMILES string and structure of the target molecule, generated with RDKit.}
\label{fig:hivsmileprompt}
\end{figure*}

\begin{figure*}[h]
\centerline{\includegraphics[width=0.8\textwidth]{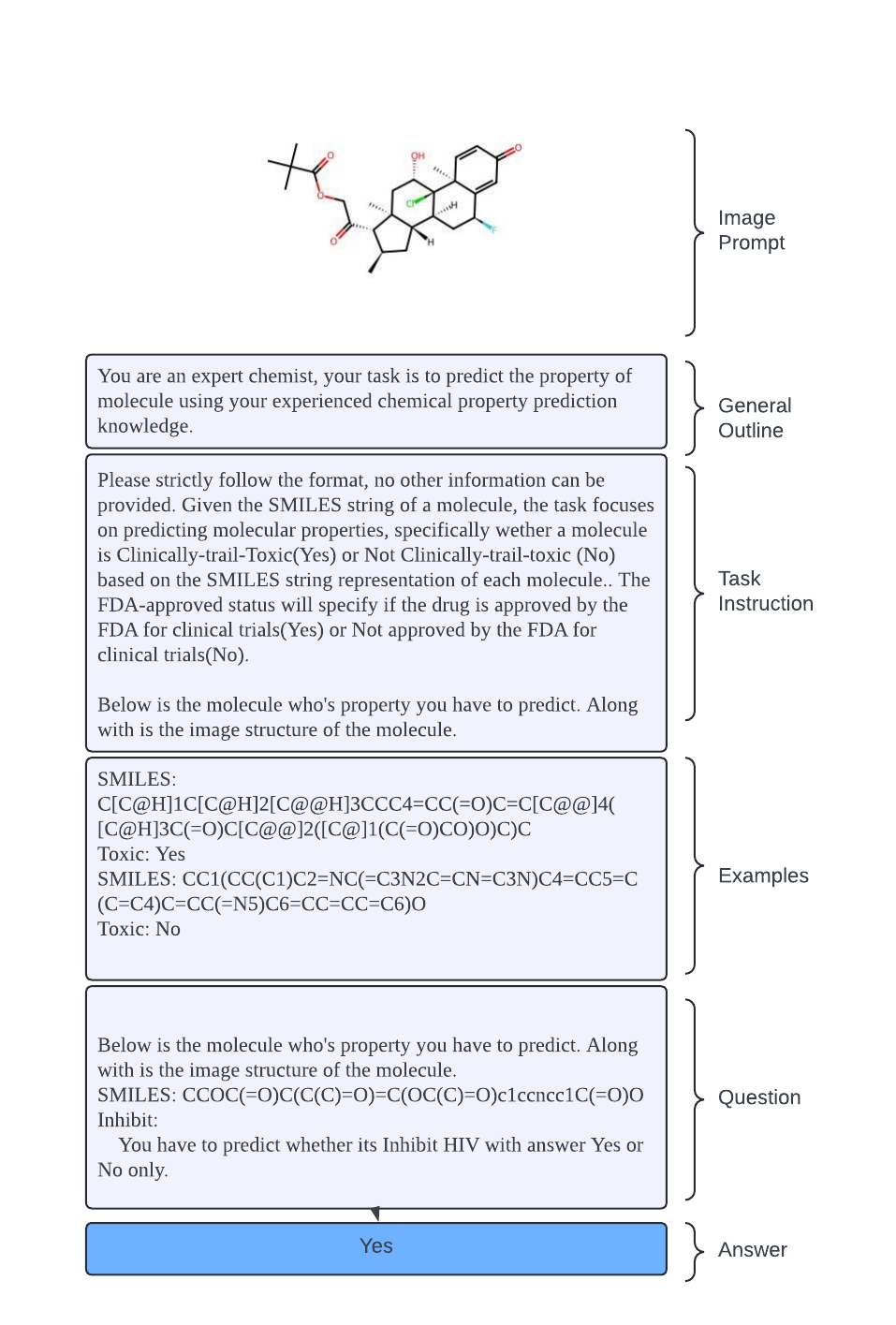}}
\caption{\textbf{\textit{Example prompt:}} The figure outlines a task for predicting whether molecules are clinically trial-toxic (ClinTox-V dataset), using their SMILES string representation. The general outline and specific instructions require a binary output (Yes/No) to indicate if the molecule is approved by the FDA for clinical trials. Example molecules are provided to illustrate the task, followed by the main question, which includes the SMILES string and structure of the target molecule, generated with RDKit.}
\label{fig:Clintoxsmileprompt}
\end{figure*}

\begin{figure*}[h]
\centerline{\includegraphics[width=0.8\textwidth]{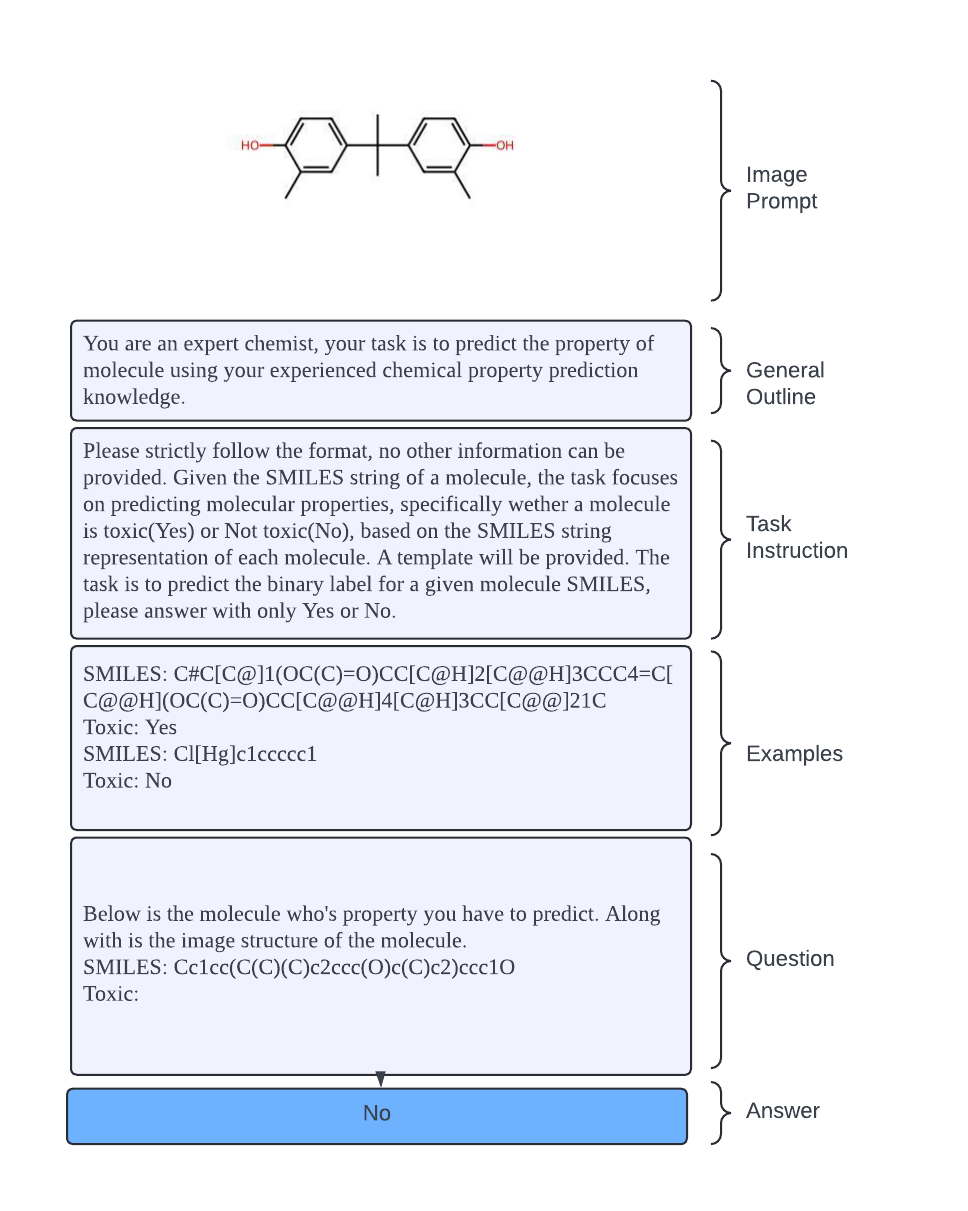}}
\caption{\textbf{\textit{Example prompt:}} The figure outlines a task for predicting the toxicity of molecules based on their SMILES string representation, specifically in the context of the Tox21 dataset. The general outline and specific instructions require a binary output (Yes/No) to indicate the molecule's toxicity. Example molecules are provided to illustrate the task, followed by the main question, which includes the SMILES string and structure of the target molecule, generated with RDKit.}
\label{fig:tox21smileprompt}
\end{figure*}

\begin{figure*}[h]
\centerline{\includegraphics[width=0.8\textwidth]{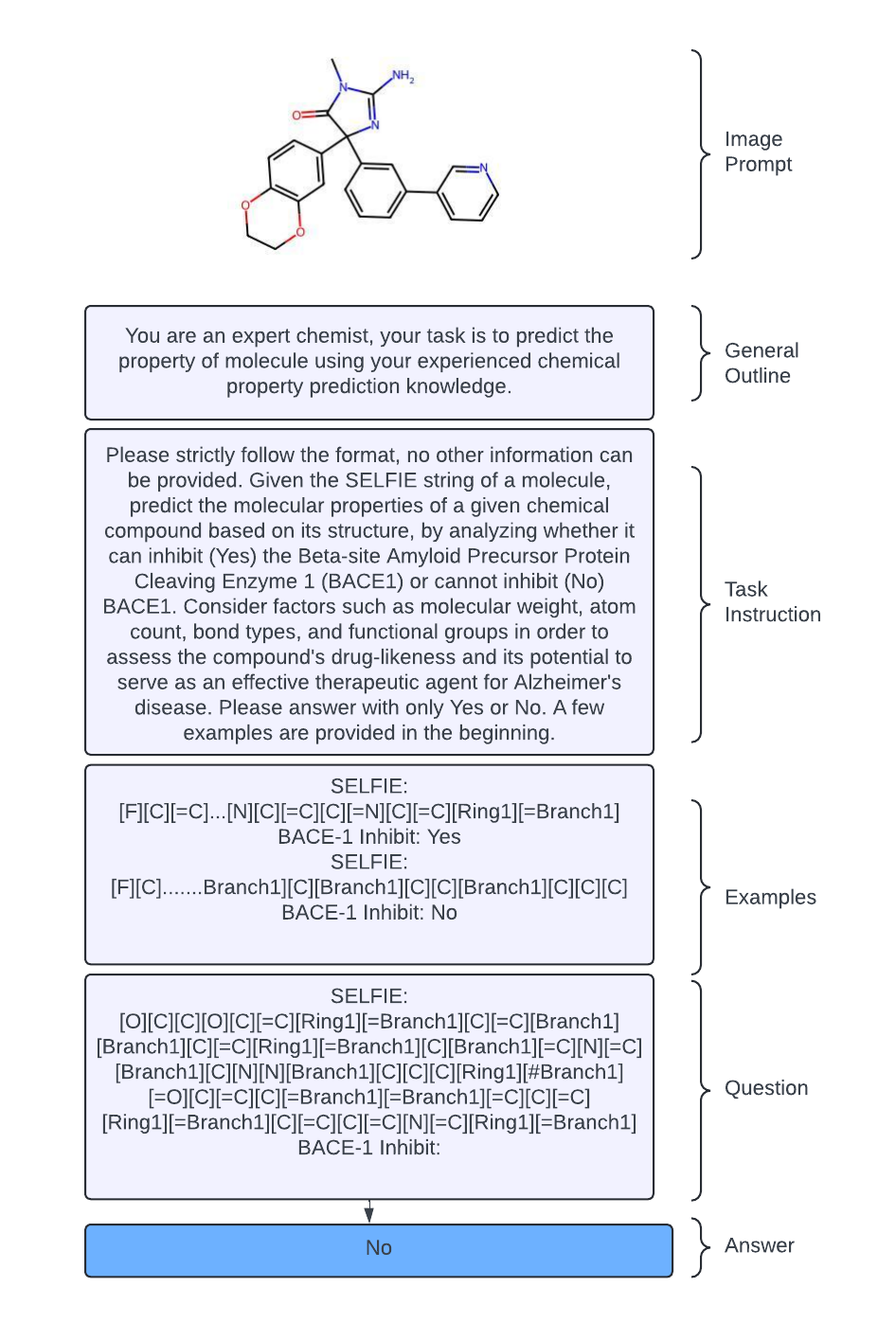}}
\caption{\textbf{\textit{Example prompt:}} The figure outlines a task for predicting the ability of molecules for BACE-Inhibit (BACE-V dataset), using their SELFIES string representation. The general outline and specific instructions require a binary output (Yes/No). Example molecules are provided to illustrate the task, followed by the main question, which includes the SELFIES string and structure of the target molecule, generated with RDKit.}
\label{fig:baceselfieprompt}
\end{figure*}

\begin{figure*}[h]
\centerline{\includegraphics[width=0.68\textwidth]{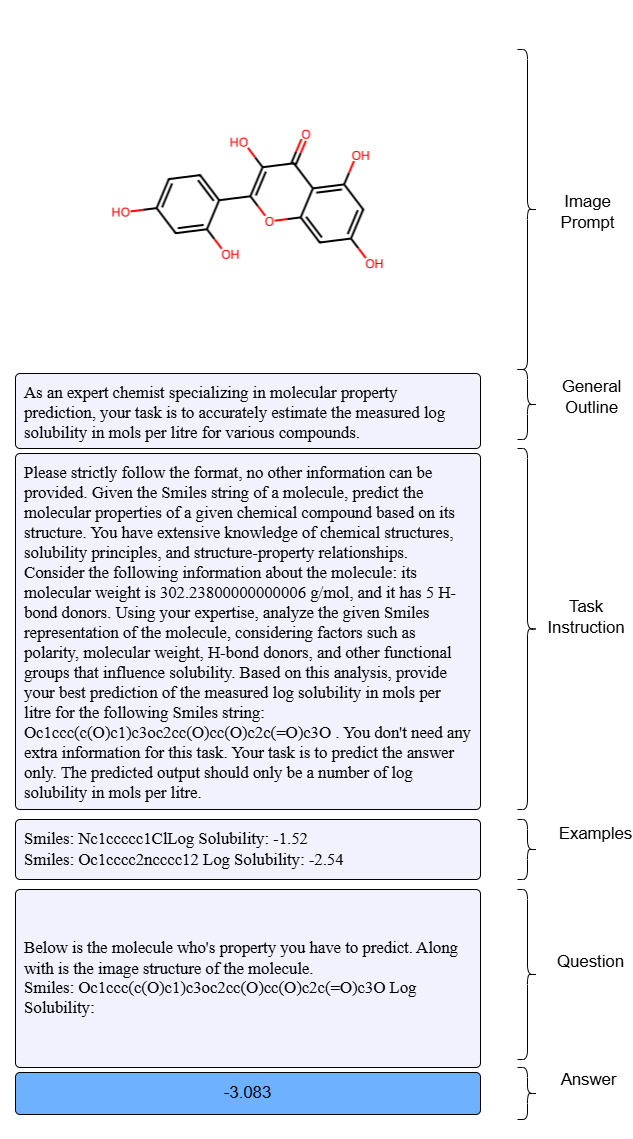}}
\caption{\textbf{\textit{Example prompt:}} The figure outlines a task for predicting the log solubility of molecules based on their SMILES string representation, using the ESOL dataset. The general outline and specific instructions require a numerical output for log solubility in mols per litre. Two example molecules (k=2) are provided to illustrate the task, followed by the main question, which includes the SMILES string and structure of the target molecule, generated with RDKit.}
\label{fig:regression_prompt}
\end{figure*}

In this section, we show some prompt examples as used for various datasets. We have included some example ICL prompts specific to some of the dataset (Figure \ref{fig:bacesmileprompt}, \ref{fig:Clintoxsmileprompt}, \ref{fig:bbbpsmileprompt}, \ref{fig:tox21smileprompt},\ref{fig:hivsmileprompt}).  
For regression tasks, we demonstrate an example using the ESOL dataset for solubility prediction (Figure \ref{fig:regression_prompt}). With ICL k=0 (different from zero-shot), we have all other section as shown in the prompt, however the example block is not used as input. The additional information as available with task instruction differentiates it from zero-shot, as models do not see `Task instruction' in zero-shot.
Prompt examples for ICL are included with SELFIES representations (Figure \ref{fig:baceselfieprompt}). 

\begin{figure*}[h]
\centerline{\includegraphics[width=0.66\textwidth]{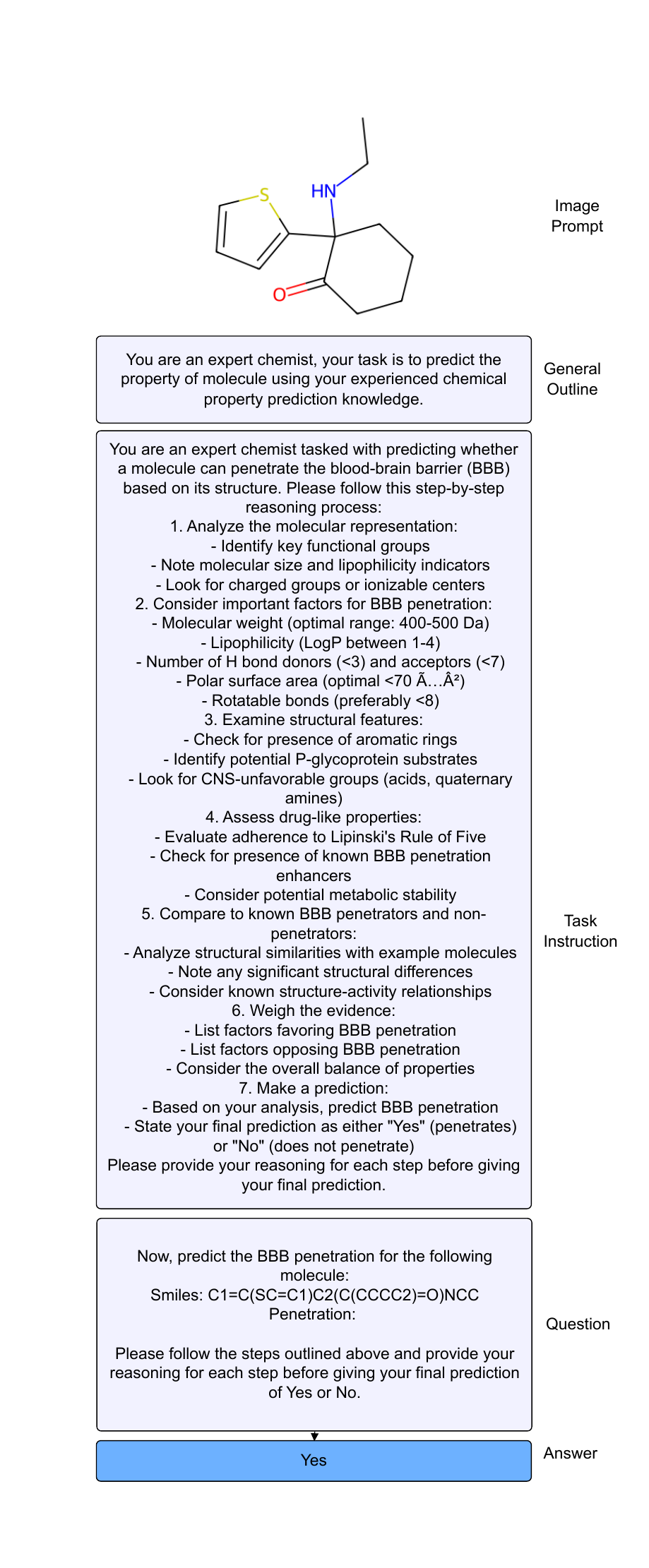}}
\caption{\textbf{\textit{Example CoT prompt:}} The figure outlines a task for predicting the Brain Penetration of molecules based on their SMILES string representation, using the BBBP dataset. The general outline and specific instructions requires a binary output (Yes/No).}
\label{fig:cot_prompt}
\end{figure*}

We include a chain-of-thought prompt example for the BBBP dataset (Figure \ref{fig:cot_prompt}), showing the step-by-step reasoning approach for classification tasks.

\begin{figure*}[h]
\centerline{\includegraphics[width=0.66\textwidth]{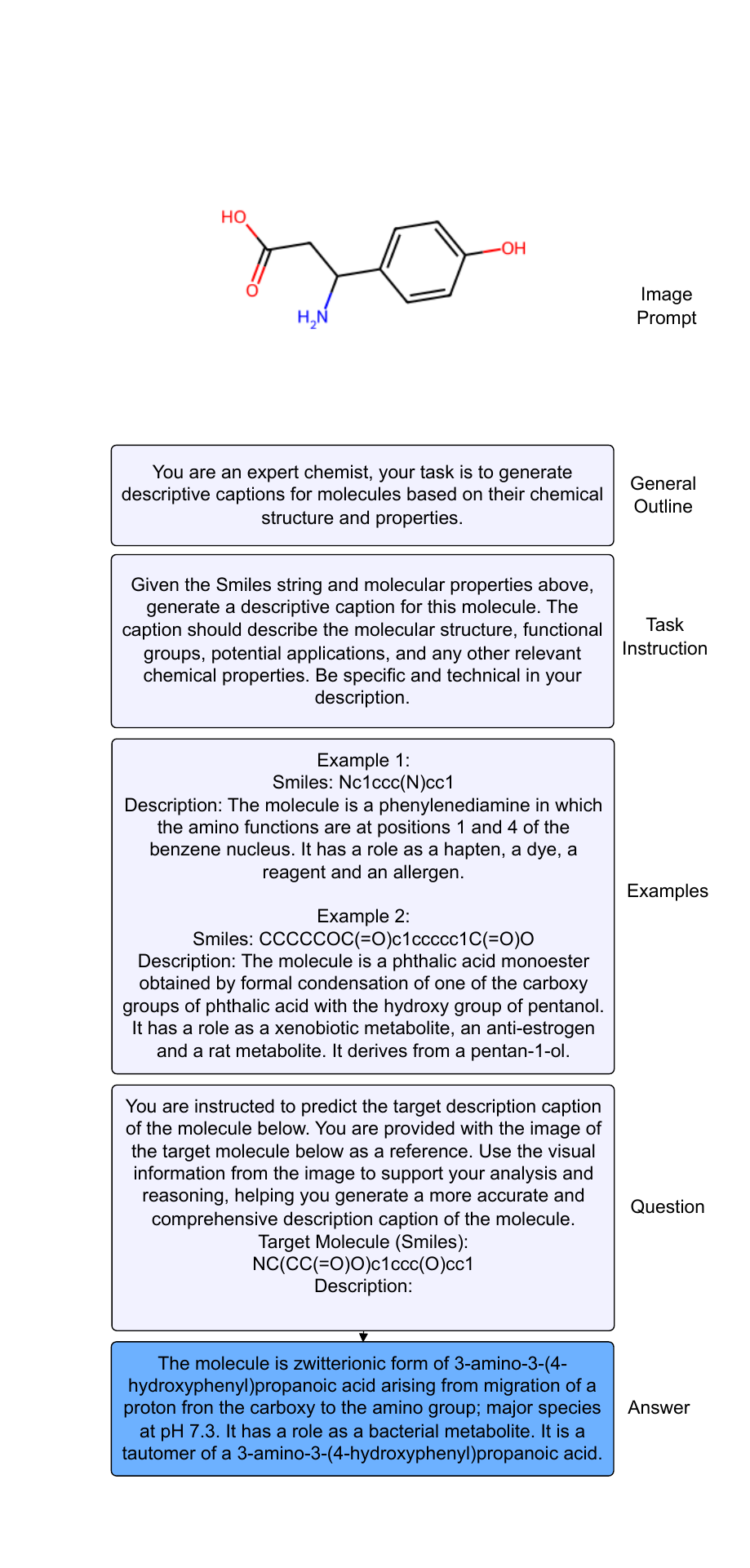}}
\caption{\textbf{\textit{Example Molecular Description prompt:}} The figure outlines a task for predicting the Molecular Description based on their SMILES string representation, using the Chebi dataset. The general outline and specific instructions requires a captioning output.}
\label{fig:molecular_description_prompt}
\end{figure*}

We provide a prompt example for molecular description tasks using the ChEBI dataset with ICL k=2 (Figure \ref{fig:molecular_description_prompt}), demonstrating how the model generates natural language descriptions of molecular structures and properties.

\end{document}